\documentclass[letterpaper]{article} 
\usepackage[preprint]{aaai2027}
\usepackage[hyphens]{url}  
\usepackage{graphicx} 
\usepackage{natbib}  
\usepackage{caption} 
\usepackage{algorithm}
\usepackage{algorithmic}
\usepackage{booktabs}
\usepackage{colortbl}
\usepackage{multirow}
\usepackage{amsmath}
\usepackage{amssymb}
\newcommand{\suppformatbreak}{\newpage}
\usepackage{tikz}
\usepackage{float}
\usepackage{afterpage}
\usepackage{pgfplots}
\pgfplotsset{compat=1.17}
\usetikzlibrary{calc}
\usepgfplotslibrary{groupplots}

\definecolor{cInk}{RGB}{60,64,72}
\definecolor{cBlue}{RGB}{0,102,162}
\definecolor{cVermil}{RGB}{204,85,0}
\definecolor{cGreen}{RGB}{0,146,110}
\definecolor{cSky}{RGB}{86,160,211}
\definecolor{cVanilla}{HTML}{999999}
\definecolor{cSingle}{HTML}{2683e0}
\definecolor{cCascade}{HTML}{4addde}
\definecolor{cQwen}{HTML}{615ced}
\definecolor{cLlama}{HTML}{66ccff}
\definecolor{cCertLine}{HTML}{EE0000}
\definecolor{cCertPoint}{HTML}{333333}
\definecolor{cFigOneBlue}{HTML}{d5f0ff}
\definecolor{cFigOnePurple}{HTML}{e4e8fb}
\definecolor{cFigOneCyan}{HTML}{c8ebf4}
\definecolor{cFigOneOutput}{HTML}{9ef6f5}
\definecolor{cTableCascade}{HTML}{D4F0FF}
\pgfplotsset{
  paper/.style={
    tick label style={font=\scriptsize},
    label style={font=\scriptsize, color=cInk},
    title style={font=\scriptsize\bfseries, color=cInk, yshift=-0.6ex},
    legend style={font=\scriptsize, draw=none, fill=none},
    legend cell align=left,
    axis line style={cInk!60, line width=0.5pt},
    tick style={cInk!60, line width=0.5pt},
    major tick length=2.2pt,
    ymajorgrids, grid style={cInk!13, line width=0.4pt},
  },
  lineP/.style={cCascade, line width=1pt, mark=*, mark size=1.7pt,
    mark options={fill=cCascade, draw=white, line width=0.5pt}},
  lineS/.style={cSingle, line width=1pt, mark=square*, mark size=1.55pt,
    mark options={fill=cSingle, draw=white, line width=0.5pt}},
  lineU/.style={cVanilla, line width=1pt, mark=triangle*, mark size=2pt,
    mark options={fill=cVanilla, draw=white, line width=0.5pt}},
  lineBase/.style={cVanilla, line width=1pt, mark=square*, mark size=1.55pt,
    mark options={fill=cVanilla, draw=white, line width=0.5pt}},
  modelL/.style={cLlama, line width=1pt, mark=*, mark size=1.7pt,
    mark options={fill=cLlama, draw=white, line width=0.5pt}},
  modelQ/.style={cQwen, line width=1pt, mark=diamond*, mark size=2.1pt,
    mark options={fill=cQwen, draw=white, line width=0.5pt}},
  refline/.style={cInk!50, dashed, line width=0.5pt},
}

\title{Doomed from the Start: Early Abort of LLM Agent Episodes via a Recall-Controlled Probe Cascade}
\author{
    Kai Ruan\textsuperscript{\rm 1},
    Zihe Huang\textsuperscript{\rm 2},
    Ziqi Zhou\textsuperscript{\rm 3},
    Qianshan Wei\textsuperscript{\rm 4},
    Jinghao Lin\textsuperscript{\rm 5},
    Xuan Wang\textsuperscript{\rm 6},
    Hao Sun\textsuperscript{\rm 1}\corresponding
}
\affiliations{
    \textsuperscript{\rm 1}Gaoling School of Artificial Intelligence, Renmin University of China\\
    \textsuperscript{\rm 2}Institute of Computing Technology, Chinese Academy of Sciences\\
    \textsuperscript{\rm 3}Duke University\\
    \textsuperscript{\rm 4}Institute of Automation, Chinese Academy of Sciences\\
    \textsuperscript{\rm 5}Independent Researcher\\
    \textsuperscript{\rm 6}College of Computer Science, Zhejiang University
}

\begin{document}

\maketitle

\begin{abstract}
Large language model (LLM) agents often waste inference compute by
continuing multi-step trajectories that are already doomed to fail. We
study early failure prediction and inference-time early stopping for
LLM agents using hidden-state probes. Lightweight linear probes on
internal activations predict eventual task failure from the first
interaction round, substantially earlier than agent-monitoring methods
based only on observable behavior.
We turn this signal into a recall-controlled abort cascade for reducing
LLM agent inference costs. The cascade applies a distribution-free
calibrated failure detector at each early interaction round and jointly
optimizes per-round recall budgets. This design ensures that eventually
successful episodes survive all early-stopping gates at a user-specified
global recall rate. After selection, the cascade is frozen and certified
on independent data, providing an exact post-selection recall guarantee.
We evaluate the method on TextCraft and WebShop with Qwen-2.5-7B,
Llama-3.2-3B, and Qwen3-1.7B. The proposed LLM agent early-stopping
cascade outperforms the best single-gate baseline in every
model-environment pair, saving $1.5$--$8.8\times$ more compute at a 90\%
recall target. Achieved recall remains within one standard deviation of
its target in all 24 configurations. The strongest settings reduce
generated tokens by $60.2\%$ on TextCraft and $54.9\%$ on WebShop at
90\% recall, while retaining savings of $45.0\%$ and $41.5\%$ at 95\%
recall. Behavior-only monitoring is consistently weaker, and adding
behavioral features to hidden-state probes provides no further gain. We
also characterize the sample complexity required to certify high-recall
early-stopping policies. The code will be released soon.
\end{abstract}

\section{Introduction}

\begin{figure}[t]
  \centering
  \begin{tikzpicture}[font=\scriptsize, >=stealth, thick,
    bx/.style={draw=cInk!60, rounded corners=1.5pt, inner sep=2.8pt, minimum height=0.48cm}]
  \node[text=cInk!75] at (3.1,0.55) {episode continues while every gate passes};
  \node[anchor=east, inner sep=0pt] at (-0.50,0)
    {\includegraphics[width=0.35cm]{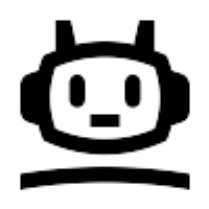}};
  \node[bx, fill=cInk!8] (r1) at (0,0) {round $1$};
  \node[bx, fill=cInk!8] (r2) at (1.8,0) {round $2$};
  \node[bx, fill=cInk!8] (r3) at (3.6,0) {round $3$};
  \node (dd) at (4.8,0) {$\cdots$};
  \node[bx, fill=cFigOneOutput] (y) at (6.2,0) {$y \in \{0,1\}$};
  \draw[->] (r1) -- (r2);
  \draw[->] (r2) -- (r3);
  \draw[->] (r3) -- (dd);
  \draw[->] (dd) -- (y);
  \foreach \i in {1,2,3} {
    \node[bx, fill=cFigOneBlue] (g\i) at ({(\i-1)*1.8},-1.05) {$f_\i(h_\i) > \tau_\i$?};
    \draw[->, cBlue] (r\i) -- (g\i)
      node[midway, right, inner sep=1.5pt, text=black] {$h_\i$};
  }
  \node[bx, fill=cFigOnePurple, minimum width=6.1cm] (ab) at (2.65,-2.05)
    {\textbf{abort}: stop episode, save remaining compute $C - c_r$};
  \foreach \i in {1,2,3} {
    \draw[->, cVermil!90!black] (g\i) -- (g\i |- ab.north);
  }
  \node[text=cVermil!90!black, inner sep=1pt, anchor=west] at (0.1,-1.55) {yes};
  \node[bx, fill=cInk!5, text width=7.3cm, align=center] at (2.9,-3.05)
    {\begin{tabular}{@{}c@{}}
     $\tau_r$: Clopper--Pearson-calibrated so gate $r$ passes $\ge t_r$ of successes\\
     $(t_1,\dots,t_{R_g})$: searched to maximize savings\\
     subject to \emph{global} recall $\ge \rho^\star$
     \end{tabular}};
  \end{tikzpicture}
  \caption{Recall-controlled abort cascade. At each of the first
  $R_g$ rounds, a linear probe $f_r$ reads the agent's hidden state
  $h_r$ and a gate aborts the episode if the failure score exceeds a
  calibrated threshold $\tau_r$; an episode completes only if it
  survives every gate. Thresholds carry per-round distribution-free
  recall guarantees, and the per-round budgets $t_r$ are jointly
  searched so that the \emph{episode-level} success recall meets a
  user-chosen target $\rho^\star$; the selected cascade can then be
  certified on independent held-out data, yielding an exact
  post-selection guarantee.}
  \label{fig:overview}
\end{figure}
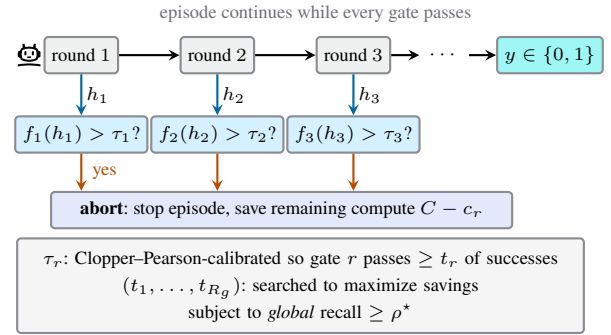

LLM-based agents increasingly tackle long-horizon tasks such as web
navigation, tool use, and embodied simulation, where a single episode
spans many rounds of interaction and consumes thousands of generated
tokens. A large fraction of this compute is wasted: when an agent has
misunderstood the task, entered an unrecoverable state, or begun to
loop, the episode is already doomed long before it formally times out
or returns a wrong answer. If we could detect such doomed episodes
early and abort them, the saved compute could be reallocated to
retries, sampling additional trajectories, or simply reducing serving
cost.

Three obstacles stand in the way. First, we need a signal that
distinguishes doomed episodes from eventually-successful ones
\emph{early} in the trajectory, and early is precisely when
behavioral evidence is scarcest. We find that scorers reading only
the agent's observable behavior are barely better than chance in the
first round and become informative only around rounds 3--4, by which
time over a third of episodes have already finished and much of the
useful remaining compute is gone. Lightweight probes on the agent's
internal activations show the opposite pattern: at the very first round
they already match or exceed the surface scorer's eventual peak,
attained only two to three rounds later, and they reach their own
peak at round 2 (Figure~\ref{fig:auc}).

Second, any abort policy is only usable in deployment if it comes with a
controllable guarantee on the harm it causes: aborting an episode
that would have succeeded silently destroys task reward, so
practitioners need a predeployment bound on the rate of such false
aborts. Third, because
agent episodes are sequential, a monitor that re-evaluates the episode
at every round faces accumulating risk. Even
if each individual check rarely kills a good episode, a successful
trajectory must survive \emph{all} of them, so per-round guarantees do
not compose into the episode-level guarantee that matters.
For example, if two gates each pass 98\% of successful episodes, the
cascade passes 98\% when they reject the same successes but only 96\%
when their false-abort sets are disjoint. Marginal per-round rates
therefore do not determine global recall.

We address these with a recall-controlled \emph{cascade}, shown in
Figure~\ref{fig:overview}. At each of an episode's first $R_g$ rounds,
a gate aborts the episode if its
probe score exceeds a threshold $\tau_r$. Each threshold is
calibrated so that an exact binomial (Clopper--Pearson) lower
confidence bound on the gate's survival rate for successful episodes
meets a per-round recall budget $t_r$, and the budget vector
$(t_1, \dots, t_{R_g})$ is searched on a disjoint validation split to
maximize compute savings subject to the \emph{global} recall (the
fraction of eventually-successful episodes that survive every
gate) meeting a user-chosen target. A safety margin makes the search
robust to validation noise. When a formal global guarantee is
required, the selected cascade is frozen and evaluated once on an
independent certification sample; because the cascade is fixed before
these data are seen, a single Clopper--Pearson lower bound on that
sample yields an exact, distribution-free guarantee that is immune to
the preceding search, no matter how many candidates the search
considered.

The cascade is strongly compute-positive across the entire evaluation
matrix. At a 90\% global recall target, the strongest TextCraft and
WebShop cells save $60.2\%$ and $54.9\%$ of generated tokens,
$1.5\times$ and $6.6\times$ the best \emph{single-gate} policy at the
same target; across all six cells the cascade's advantage over the
best single gate ranges from $1.5\times$ to $8.8\times$, confirming
that much of the practical value comes from distributing recall budget
across rounds. At a conservative 95\% target the cascade still
saves up to $45.0\%$ (TextCraft) and $41.5\%$ (WebShop), and its
achieved recall stays within one standard deviation of its target in
all 24 evaluated configurations. Just as important, when the signal is
weak, the cascade degrades gracefully toward a conservative near-no-op
policy that prioritizes successful episodes, the appropriate failure
mode for a deployed monitor. Holding the scorer fixed, the allocation
comparison shows that recall budgets should track when each cell
becomes predictable; a separate ablation attributes the early signal itself to
the activations: behavior-only monitoring is consistently weaker, and
stacking behavioral features onto the probe adds nothing.

Finally, we give an honest account of what \emph{certified} recall
control costs in data. With $n$ successful episodes in an independent
certification sample, a one-sided certificate at confidence $95\%$
can support recall targets only up to $0.05^{1/n}$: $114$ successes
cap certifiable targets near $0.974$, while targets of $0.98$ and
$0.99$ require roughly $149$ and $299$ successful episodes
(Figure~\ref{fig:cert}). The certificate makes this data requirement
explicit: the same machinery that saves compute also tells the
practitioner, before deployment, which recall promises the available
data support.

Our contributions are:
\begin{itemize}
\item We demonstrate across a full $2\times3$ matrix that eventual
  task failure of an LLM agent is predictable
  from internal activations within the first interaction rounds,
  before low-cost behavioral features become comparably informative.
\item We propose, to our knowledge, the first abort policy for LLM
  agents that targets \emph{episode-level} success recall across
  multiple sequential decision points, via per-round
  distribution-free gates whose recall budgets are jointly optimized
  under a global constraint, together with an exact,
  distribution-free post-selection certificate on independent data.
\item We show the cascade saves up to $60.2\%$ of inference compute
  at 90\% global recall, outperforming the best single-gate policy in
  every cell, characterize the data cost of certifying stricter
  targets, and use ALFWorld as a stress test showing that the method
  fails conservatively and preserves recall when the signal is weak.
\end{itemize}

\newcommand{\relatedworksection}{%
\section{Related Work}

\paragraph{Internal representations.}
Activation probes recover latent knowledge, truth, hallucination, and
outcome signals
\citep{kadavath2022language,burns2023discovering,marks2023geometry,
azaria2023internal,orgad2025llms,afzal2025knowing,zhang2025reasoning}.
Agent representations also expose success/failure directions
\citep{anon2026conformal_interp}; we turn this early signal into a
stopping rule with controlled retained-success recall.

\paragraph{Agent monitoring.}
Agent monitors use text auditors, weak supervision, runtime statistics,
prefix warnings, and structured traces
\citep{zhang2026foresight,anon2026alphastop,anon2026observability,
barke2026agentrx,chen2026signals,pham2026agentstop,
mittapalli2026trace,dellibarda2026poirot,huang2026prefixguard,
zhao2026grade}. Our monitor reads frozen-policy
activations without an auxiliary LLM pass and jointly calibrates recall
over the \emph{entire sequence} of abort decisions.

\paragraph{Risk control.}
Distribution-free calibration provides finite-sample guarantees under
exchangeability
\citep{angelopoulos2023gentle,bates2021distribution,
angelopoulos2024conformal,angelopoulos2025learn}. CALM controls sequence-level token exits,
FIPER calibrates rollout alarms, and dynamic abstention learns when
reasoning should quit
\citep{schuster2022confident,romer2025fiper,davidov2026quit}. Our
setting controls episode-level retained-success recall across agent
gates and independently certifies the selected cascade.

\paragraph{Adaptive compute.}
Adaptive inference methods stop self-consistency, route between models,
predict ongoing success, prune partial trajectories, or end futile
reasoning
\citep{aggarwal2023lets,li2024escape,manvi2024adaptive,
chen2023frugalgpt,zellinger2025cost,kim2026atropos,lightman2024lets,
guan2026futile,jiang2025runaway,shrivastava2026semantic,
anon2026bagen,anon2025refrain,mao2025escot}. We decide whether a
running agent episode is worth finishing and pair saved tokens with
calibrated success recall.

}%

\section{Method}

\begin{algorithm}[t]
\caption{Recall-Controlled Abort Cascade with Optional Certification}
\label{alg:cascade}
\textbf{Input}: labeled episodes $\mathcal{D}$, global recall target $\rho^\star$, budget grid $\mathcal{T}$, margin $\delta$; optional independent certification set $\mathcal{D}_{\mathrm{cert}}$ and level $\alpha_m$\\
\textbf{Output}: gates $\{(f_r, \tau_r)\}_{r=1}^{R_g}$
\begin{algorithmic}[1]
\STATE Score all episodes by task-grouped cross-fitted probes $f_r$
\STATE Partition tasks into calibration / search-validation / test; keep $\mathcal{D}_{\mathrm{cert}}$ independent if supplied
\FOR{each candidate budget $\mathbf{t} \in \mathcal{T}$}
\FOR{$r = 1, \dots, R_g$}
\STATE $\tau_r(\mathbf{t}) \leftarrow$ smallest threshold whose Clopper--Pearson lower bound on per-round survival of successful calibration episodes at $r$ is $\ge t_r$
\ENDFOR
\STATE Simulate cascade on validation split; record global recall $\hat{\rho}_{\mathrm{val}}$ and savings
\ENDFOR
\STATE $\mathcal{F} \leftarrow$ candidates with $\hat{\rho}_{\mathrm{val}} \ge \rho^\star + \delta$
\IF{$\mathcal{F} = \emptyset$}
\STATE Abstain (no aborts)
\ELSE
\STATE Deploy $\mathbf{t}^\star = \arg\max_{\mathbf{t} \in \mathcal{F}}$ validation savings
\ENDIF
\IF{$\mathcal{D}_{\mathrm{cert}}$ is supplied and its global Clopper--Pearson lower bound for $\mathbf{t}^\star$ is $<\rho^\star$}
\STATE Abstain (no aborts)
\ENDIF
\STATE Evaluate once on the test split
\end{algorithmic}
\end{algorithm}

\subsection{Problem Setup}

An agent episode is a sequence of interaction rounds
$(s_1, a_1, s_2, a_2, \dots)$ between an LLM policy and an
environment, terminating with a binary outcome $y \in \{0, 1\}$. We
index rounds from one; episodes run up to $R_{\mathrm{full}} = 20$
rounds. Let $c_r$ denote the cumulative inference cost incurred
through round $r$ and $C$ the total cost of the full episode.

An \emph{abort cascade} places a gate at each of the first $R_g$
rounds. We use $R_g = 6$ throughout. At gate $r$, every
episode still running is scored by a per-round scorer $f_r$ and
aborted if $f_r(x) > \tau_r$, losing any potential success but
saving the remaining cost $C - c_r$. Episodes terminate at different
rounds; in our data, over a third of episodes finish within the
first two rounds (see supplementary alive-episode curves), so a late gate
sees only the episodes still alive there: late gates incur less recall
risk and guard less remaining compute. This trade-off is
exactly what the cascade optimizes.

Aborting an episode that would have succeeded is the critical error
mode. We target \emph{global success recall}: among episodes with
$y = 1$, the fraction that survive \emph{every} gate and run to
completion. The design goal is to maximize expected cost savings
subject to a user-specified global recall floor $\rho^\star$. Note
that controlling the recall of each gate in isolation does not
control the global recall: false-abort probability accumulates across
gates, and the accumulation depends on how many successes each gate
sees.

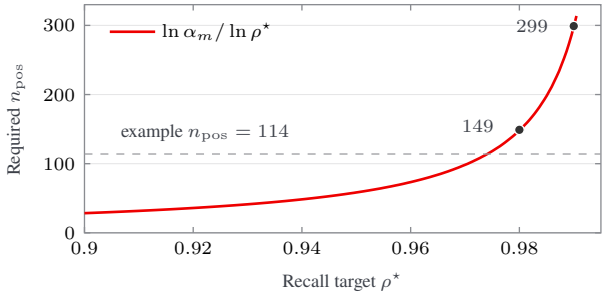
\begin{figure}[t]
  \centering
  \begin{tikzpicture}
\begin{axis}[paper,
  width=\linewidth, height=4.6cm,
  xlabel={Recall target $\rho^\star$},
  ylabel={Required $n_{\mathrm{pos}}$},
  xmin=0.90, xmax=0.995, ymin=0, ymax=330,
  xtick={0.90,0.92,0.94,0.96,0.98},
  x tick label style={font=\scriptsize, /pgf/number format/.cd, fixed, precision=2},
  legend pos=north west]
\addplot[cCertLine, line width=1pt, domain=0.90:0.9905, samples=200] {ln(0.05)/ln(x)};
\addplot[refline] coordinates {(0.90,114)(0.995,114)};
\addplot[cCertPoint, only marks, mark=*, mark size=1.8pt,
  mark options={fill=cCertPoint, draw=white, line width=0.5pt}] coordinates
  {(0.98,149)(0.99,299)};
\node[font=\scriptsize, text=cInk, anchor=south west] at (axis cs:0.905,118)
  {example $n_{\mathrm{pos}}=114$};
\node[font=\scriptsize, text=cInk, anchor=east] at (axis cs:0.977,155) {$149$};
\node[font=\scriptsize, text=cInk, anchor=east] at (axis cs:0.987,299) {$299$};
\legend{$\ln \alpha_m / \ln \rho^\star$}
\end{axis}
\end{tikzpicture}
  \caption{Sample complexity of the global recall certificate
  ($\alpha_m = 0.05$): minimum successful certification episodes. With
  $n_{\mathrm{pos}}=114$, targets up to $\approx0.974$ are supported;
  targets $0.98$ and $0.99$ require 149 and 299 successes, independent
  of scorer quality.}
  \label{fig:cert}
\end{figure}

\subsection{Per-Round Failure Scorers}

At each gate round $r$, we extract a feature vector
$x \in \mathbb{R}^d$ from the agent LLM's internal activations:
the residual-stream hidden state at the final token of the agent's
generated action in round $r$, at a single layer selected once per
matrix cell by a preliminary per-layer probe-AUC sweep.
In our experiments this vector is recovered by a teacher-forced
forward pass over the logged trajectory. Online deployment can instead
collect selected-layer hidden states through an instrumented serving
stack during inference (see Discussion), avoiding offline replay.
Layer choice is fixed before any recall-target or cascade evaluation
and remains fixed across targets.
A per-round probe $f_r$, a logistic regression on standardized
features with $L_2$ regularization and $C = 1$, is trained to predict
eventual failure $1 - y$ from episodes alive at round $r$.

We compare against two alternative scorers, holding the entire
downstream calibration and search pipeline fixed. The
\textbf{surface} scorer observes only the agent's \emph{behavior}: a
logistic model over
trajectory features observable from the serving API alone: the mean
action-token log-probability of the current round, the mean over
preceding rounds, the number of generated tokens, the prefix length,
and the count of preceding rounds whose environment feedback contains
error keywords (``error'', ``invalid'', ``fail'', etc.).
The \textbf{stacking} scorer concatenates these surface features
onto the probe's activation features, testing whether behavioral
evidence adds anything beyond the hidden state.
The surface model is deliberately cheap and hand-engineered, matching
the negligible-overhead regime our monitor operates in; richer
structured-trace monitors are complementary, and an apples-to-apples
comparison would place them inside the same recall-calibrated cascade
while accounting for their inference overhead.

\subsection{Per-Round Recall-Calibrated Gates}

Each gate's threshold is set on a calibration split so that the gate
provably passes at least a $t_r$ fraction of successful episodes.
Given a per-round recall budget $t_r$, let
$\mathcal{S}_r = \{f_r(x_i) : y_i = 1,\ i \text{ alive at } r\}$ be
the scores of the $n_r$ successful calibration episodes alive at
round $r$. For a candidate threshold $\tau$, let
$k(\tau) = |\{s \in \mathcal{S}_r : s \le \tau\}|$ be the number of
survivors; the exact binomial (Clopper--Pearson) lower confidence
bound on the gate's true survival rate is the Beta quantile
\begin{equation}
\underline{p}(\tau) \;=\; \mathrm{Beta}^{-1}\!\left(\alpha;\; k(\tau),\; n_r - k(\tau) + 1\right),
\end{equation}
and we set $\tau_r$ to the smallest calibration score with
$\underline{p}(\tau_r) \ge t_r$, at per-gate confidence level
$1 - \alpha$ with $\alpha = 0.05$. This is deliberately stronger than a marginal conformal
quantile: the per-round guarantee holds with high confidence over the
calibration draw, at the price of conservatism when $n_r$ is small
(a gate abstains, i.e., aborts nothing, whenever $n_r$ is insufficient
to support $t_r$). A conformal-quantile variant
($\tau_r = \mathrm{Quantile}_{\lceil (n_r+1) t_r \rceil / n_r}(\mathcal{S}_r)$)
saves more but can violate its target with small calibration sets
(technical appendix, ``Conformal-Quantile Gates''); we therefore use
the high-confidence Clopper--Pearson gate throughout. A budget of
$t_r = 1$ disables the gate.

\begin{figure*}[t]
  \centering
  \begin{tikzpicture}
\begin{groupplot}[
  paper,
  group style={group size=2 by 1, horizontal sep=0.65cm,
    ylabels at=edge left, yticklabels at=edge left},
  width=0.50\textwidth, height=4.0cm,
  xmin=0.6, xmax=6.4, xtick={1,2,3,4,5,6},
  ymin=0.5, ymax=1.0, ylabel={Post-gen probe AUC}, xlabel={Gate round},
  legend columns=3]
\nextgroupplot[title={TextCraft}, legend to name=auclegend]
\addplot[modelQ] coordinates {(1,0.860)(2,0.895)(3,0.897)(4,0.889)(5,0.888)(6,0.848)};
\addplot[modelL] coordinates {(1,0.698)(2,0.738)(3,0.692)(4,0.648)(5,0.632)(6,0.588)};
\addplot[cGreen, line width=1pt, mark=triangle*, mark size=2pt,
  mark options={fill=cGreen, draw=white, line width=0.5pt}] coordinates
  {(1,0.811)(2,0.895)(3,0.843)(4,0.765)(5,0.762)(6,0.758)};
\addplot[refline, forget plot, samples=2, domain=0.6:6.4] {0.5};
\legend{Qwen-2.5-7B, Llama-3.2-3B, Qwen3-1.7B}
\nextgroupplot[title={WebShop}]
\addplot[modelQ] coordinates {(1,0.699)(2,0.743)(3,0.761)(4,0.731)(5,0.758)(6,0.826)};
\addplot[modelL] coordinates {(1,0.523)(2,0.635)(3,0.738)(4,0.751)(5,0.797)(6,0.818)};
\addplot[cGreen, line width=1pt, mark=triangle*, mark size=2pt,
  mark options={fill=cGreen, draw=white, line width=0.5pt}] coordinates
  {(1,0.591)(2,0.829)(3,0.896)(4,0.914)(5,0.941)(6,0.922)};
\addplot[refline, forget plot, samples=2, domain=0.6:6.4] {0.5};
\end{groupplot}
\node[anchor=south] at ($(group c1r1.north)!0.5!(group c2r1.north)+(0,0.35cm)$)
  {\pgfplotslegendfromname{auclegend}};
\end{tikzpicture}
  \caption{Cross-fitted post-generation probe AUC for predicting
  eventual episode failure at each gate round in the full
  $2\times3$ matrix. The strongest transfer appears on WebShop with
  Qwen3-1.7B; WebShop Llama becomes informative only after the first
  two rounds, matching the later-gate savings in Table~\ref{tab:main}.}
  \label{fig:auc}
\end{figure*}
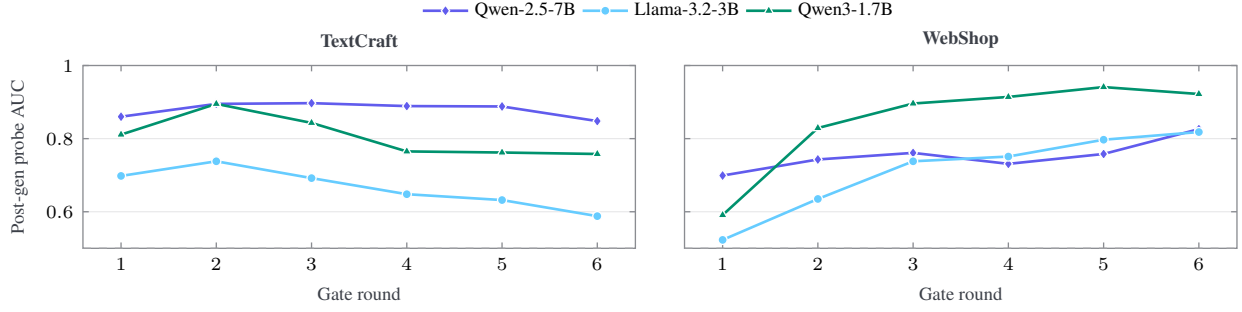

\afterpage{%
\begin{table*}[t]
\centering
\scriptsize
\setlength{\tabcolsep}{3.5pt}
\begin{tabular}{@{}lllcccc@{}}
\toprule
Environment & Model & Allocation & $0.90$ & $0.92$ & $0.95$ & $0.97$ \\
\midrule
\multirow{3}{*}{TextCraft} & \multirow{3}{*}{Qwen-2.5-7B} & Uniform & $8.3 \pm 4.8$ & $8.3 \pm 4.8$ & $0.0 \pm 0.0$ & $0.0 \pm 0.0$ \\
 & & Single & $40.1 \pm 7.1$ & $34.8 \pm 7.8$ & $25.8 \pm 9.0$ & $15.9 \pm 10.3$ \\
 & & \cellcolor{cTableCascade}Cascade & \cellcolor{cTableCascade}$\mathbf{60.2 \pm 6.8}$ & \cellcolor{cTableCascade}$\mathbf{56.6 \pm 10.1}$ & \cellcolor{cTableCascade}$\mathbf{45.0 \pm 12.5}$ & \cellcolor{cTableCascade}$\mathbf{31.6 \pm 15.0}$ \\
\multirow{3}{*}{TextCraft} & \multirow{3}{*}{Llama-3.2-3B} & Uniform & $3.4 \pm 2.8$ & $0.0 \pm 0.0$ & $0.0 \pm 0.0$ & $0.0 \pm 0.0$ \\
 & & Single & $22.0 \pm 6.4$ & $17.1 \pm 4.8$ & $10.7 \pm 3.8$ & $6.0 \pm 3.4$ \\
 & & \cellcolor{cTableCascade}Cascade & \cellcolor{cTableCascade}$\mathbf{34.7 \pm 9.2}$ & \cellcolor{cTableCascade}$\mathbf{25.8 \pm 11.0}$ & \cellcolor{cTableCascade}$\mathbf{17.2 \pm 7.2}$ & \cellcolor{cTableCascade}$\mathbf{6.9 \pm 7.1}$ \\
\multirow{3}{*}{TextCraft} & \multirow{3}{*}{Qwen3-1.7B} & Uniform & $5.9 \pm 6.0$ & $4.6 \pm 4.5$ & $0.0 \pm 0.0$ & $0.0 \pm 0.0$ \\
 & & Single & $29.9 \pm 8.6$ & $23.2 \pm 8.5$ & $14.3 \pm 7.9$ & $7.1 \pm 5.1$ \\
 & & \cellcolor{cTableCascade}Cascade & \cellcolor{cTableCascade}$\mathbf{47.3 \pm 12.2}$ & \cellcolor{cTableCascade}$\mathbf{43.3 \pm 12.9}$ & \cellcolor{cTableCascade}$\mathbf{26.8 \pm 13.0}$ & \cellcolor{cTableCascade}$\mathbf{14.8 \pm 12.7}$ \\
\midrule
\multirow{3}{*}{WebShop} & \multirow{3}{*}{Qwen-2.5-7B} & Uniform & $0.0 \pm 0.0$ & $0.0 \pm 0.0$ & $0.0 \pm 0.0$ & $0.0 \pm 0.0$ \\
 & & Single & $19.2 \pm 6.9$ & $14.4 \pm 5.2$ & $8.0 \pm 5.2$ & $5.0 \pm 4.2$ \\
 & & \cellcolor{cTableCascade}Cascade & \cellcolor{cTableCascade}$\mathbf{35.9 \pm 5.7}$ & \cellcolor{cTableCascade}$\mathbf{31.8 \pm 5.4}$ & \cellcolor{cTableCascade}$\mathbf{20.6 \pm 8.2}$ & \cellcolor{cTableCascade}$\mathbf{11.3 \pm 7.8}$ \\
\multirow{3}{*}{WebShop} & \multirow{3}{*}{Llama-3.2-3B} & Uniform & $0.0 \pm 0.0$ & $0.0 \pm 0.0$ & $0.0 \pm 0.0$ & $0.0 \pm 0.0$ \\
 & & Single & $2.8 \pm 1.8$ & $2.1 \pm 1.3$ & $1.1 \pm 1.1$ & $0.0 \pm 0.0$ \\
 & & \cellcolor{cTableCascade}Cascade & \cellcolor{cTableCascade}$\mathbf{24.7 \pm 8.4}$ & \cellcolor{cTableCascade}$\mathbf{20.0 \pm 10.1}$ & \cellcolor{cTableCascade}$\mathbf{13.6 \pm 9.6}$ & \cellcolor{cTableCascade}$\mathbf{12.0 \pm 10.2}$ \\
\multirow{3}{*}{WebShop} & \multirow{3}{*}{Qwen3-1.7B} & Uniform & $0.0 \pm 0.0$ & $0.0 \pm 0.0$ & $0.0 \pm 0.0$ & $0.0 \pm 0.0$ \\
 & & Single & $8.3 \pm 4.8$ & $6.1 \pm 4.4$ & $3.8 \pm 4.2$ & $1.8 \pm 1.3$ \\
 & & \cellcolor{cTableCascade}Cascade & \cellcolor{cTableCascade}$\mathbf{54.9 \pm 5.7}$ & \cellcolor{cTableCascade}$\mathbf{52.3 \pm 6.6}$ & \cellcolor{cTableCascade}$\mathbf{41.5 \pm 15.8}$ & \cellcolor{cTableCascade}$\mathbf{26.4 \pm 17.1}$ \\
\bottomrule
\end{tabular}
\caption{Compute saved (\%) for the searched stacking cascade, best
single gate, and uniform allocation across all environments and
models. The cascade dominates both allocation baselines in all 24
configurations; achieved recall stays within one standard deviation
of its target throughout (Figure~\ref{fig:recall}).}
\label{tab:main}
\end{table*}
}
\subsection{Recall Budget Search Under a Global Constraint}

Per-round guarantees do not compose multiplicatively in any useful
way: a union bound over gates remains valid yet is overly conservative,
because it ignores that late gates expose few successes and that
per-gate false-abort events are far from disjoint. We therefore treat
the budget vector $\mathbf{t} = (t_1, \dots, t_{R_g})$ as a
hyperparameter and select it empirically on a validation split
disjoint from calibration.

Concretely, probe scores are produced by task-grouped stratified
group $k$-fold cross-fitting, so every episode is scored by a probe
that never saw its task during training. Tasks are then partitioned
into a calibration set for gate thresholds ($20\%$ of tasks), a
validation set for budget search ($20\%$), and a held-out test set
($60\%$). Grouping by task throughout ensures no task contributes
episodes to two sides of any split.

For each candidate $\mathbf{t}$ on the grid
$t_r \in \{0.85, 0.90, 0.95, 0.98, 0.99, 1.0\}$ ($6^6 = 46{,}656$
candidates), we calibrate all gates on the calibration split,
simulate the full cascade on the validation split, and record its
global recall $\hat{\rho}_{\mathrm{val}}(\mathbf{t})$ and compute
savings. The deployed budget maximizes validation savings subject to
a feasibility condition, in a two-stage design that separates
flexible search from statistically clean certification:
\begin{itemize}
\item \textbf{Margin} (default): require
  $\hat{\rho}_{\mathrm{val}}(\mathbf{t}) \ge \rho^\star + \delta$
  with a fixed safety margin $\delta = 0.02$, a guard against
  selection bias from searching over many candidates. We size the
  margin from the binomial standard error of validation recall and
  validate its coverage empirically across the full matrix and in a
  dedicated margin-size sweep (technical appendix, ``Margin Size
  Sweep'').
\item \textbf{Independent post-selection certificate} (optional):
  after selecting $\mathbf{t}^\star$, freeze the cascade and compute a
  Clopper--Pearson lower bound at level $1-\alpha_m$
  ($\alpha_m=0.05$) on its global recall using successful episodes in
  an independent certification sample; deploy only if the bound
  exceeds $\rho^\star$. Because the cascade is fixed before these
  data are observed, the bound is exact and distribution-free
  regardless of the preceding search size. This converts the margin's empirical control into a
  formal, a priori verifiable guarantee whenever one is required,
  without constraining the search itself.
\end{itemize}
If no candidate is margin-feasible, or if an optional independent
certificate fails, the policy abstains and aborts nothing: the
method's failure mode is conservative by construction.
Two fixed budget allocations serve as structural baselines:
\textbf{single-gate}, which spends the entire recall budget at a
single round, with both the round and its budget selected by the
same validation search (recovering the single-decision-point
policies of prior monitoring work as a special case of our
framework), and
\textbf{uniform}, which sets the same $t_r$ at every gate.

\subsubsection{Sample complexity of certification.}
The certificate makes visible a fundamental data requirement. Let
$n_{\mathrm{pos}}$ be the number of successful episodes in the
independent certification sample. Even a
candidate that aborts \emph{nothing} has a lower bound on an
independent certification sample of
$\mathrm{Beta}^{-1}(\alpha_m; n_{\mathrm{pos}}, 1) =
\alpha_m^{1/n_{\mathrm{pos}}}$ (the ``rule of three''), so targets
above this level are unattainable regardless of scorer quality. For
example, around $114$ successful certification episodes support targets
up to about $0.974$, while targets $0.98$ and $0.99$ require roughly
$149$ and $299$ successful certification episodes. The resulting sample
complexity of stricter targets is analyzed in the Results
(Figure~\ref{fig:cert}) and applies to every matrix cell
through its own number of certification successes.

\section{Experimental Setup}
\subsubsection{Environment and Agents.}
We evaluate a $2\times3$ matrix: two environments and three agent
policies. TextCraft \citep{prasad2024adapt} is a text-based crafting
environment from AgentGym \citep{xi2024agentgym} in which an agent
must synthesize a target item by navigating multi-step recipes.
WebShop is an instruction-following shopping environment in which the
agent searches, clicks products and options, and buys an item matching
the user request. The three agent policies are Llama-3.2-3B
\citep{meta2024llama32}, Qwen-2.5-7B \citep{qwen2025qwen25}, and
Qwen3-1.7B. Every cell contains $800$ rollout episodes. TextCraft uses
$100$ tasks with $8$ rollouts per task; WebShop uses $200$ tasks with
$4$ rollouts per task, keeping the episode count fixed while
increasing task diversity. TextCraft episodes run up to
$R_{\mathrm{full}} = 20$ rounds, WebShop episodes run up to $10$
rounds, and gates are placed at rounds $1$--$6$ in all cells. The
supplementary alive-episode curves show that TextCraft episodes
often terminate early, while WebShop episodes remain alive longer.
Compute savings are reported as the fraction of total generated
tokens saved by aborts, counting only each aborted episode's
remaining computation.

\subsubsection{Protocol.}
Probes are per-round logistic regressions with $C = 1$ and standardized
features, applied to the fixed layer selected for each model family:
layer 14 for Llama-3.2-3B, layer 20 for Qwen-2.5-7B, and layer 28
for Qwen3-1.7B in the full-matrix runs. On one server with eight
NVIDIA H20 GPUs, the entire pipeline (task-level splitting, probe
cross-fitting, gate calibration, budget search, evaluation) is
repeated over $20$ random seeds; we report test-split mean $\pm$
standard deviation. All
scorers and all budget-allocation baselines share identical splits,
calibration machinery, and search procedure, so differences isolate
the quality of the underlying signal (scorer comparisons) or the
value of distributing the recall budget (allocation comparisons).
Unless otherwise stated, the allocation tables use the stacking
scorer (activation probe plus surface features) as the common scorer:
it contains both feature families, giving behavioral evidence every
opportunity to contribute. Figure~\ref{fig:scorer} shows the probe
alone performs equivalently, so the reported gains attribute to the
activations. For Qwen3-1.7B we use non-thinking generation mode for
action-format compatibility; for WebShop with Llama-3.2-3B we use a
format-constrained WebShop instruction to reduce invalid actions.

\afterpage{%
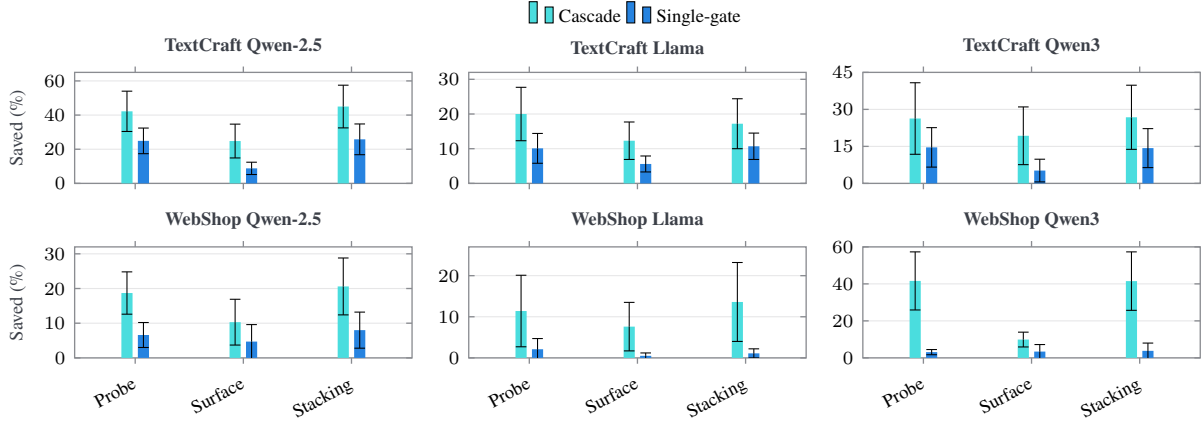
\begin{figure*}[t]
  \centering
  \begin{tikzpicture}
\begin{groupplot}[
  paper,
  group style={group size=3 by 2, horizontal sep=0.75cm, vertical sep=0.84cm,
    ylabels at=edge left, yticklabels at=all},
  width=0.34\textwidth, height=3.05cm,
  ybar, /pgf/bar width=4.2pt,
  symbolic x coords={Probe, Surface, Stacking},
  xtick=data, enlarge x limits=0.28,
  x tick label style={font=\scriptsize, rotate=28, anchor=north east},
  ylabel={Saved (\%)},
  legend columns=2]
\nextgroupplot[title={TextCraft Qwen-2.5}, ymin=0, ymax=65,
  ytick={0,20,40,60}, xticklabels=\empty,
  legend to name=scorerlegend]
\addplot[fill=cCascade, draw=none, error bars/.cd, y dir=both, y explicit] coordinates
  {(Probe,42.2) +- (0,11.8) (Surface,24.8) +- (0,9.9) (Stacking,45.0) +- (0,12.5)};
\addplot[fill=cSingle, draw=none, error bars/.cd, y dir=both, y explicit] coordinates
  {(Probe,24.9) +- (0,7.5) (Surface,8.8) +- (0,3.6) (Stacking,25.8) +- (0,9.0)};
\legend{Cascade, Single-gate}
\nextgroupplot[title={TextCraft Llama}, ymin=0, ymax=32,
  ytick={0,10,20,30}, xticklabels=\empty]
\addplot[fill=cCascade, draw=none, error bars/.cd, y dir=both, y explicit] coordinates
  {(Probe,20.0) +- (0,7.7) (Surface,12.3) +- (0,5.4) (Stacking,17.2) +- (0,7.2)};
\addplot[fill=cSingle, draw=none, error bars/.cd, y dir=both, y explicit] coordinates
  {(Probe,10.1) +- (0,4.3) (Surface,5.6) +- (0,2.3) (Stacking,10.7) +- (0,3.8)};
\nextgroupplot[title={TextCraft Qwen3}, ymin=0, ymax=45,
  ytick={0,15,30,45}, xticklabels=\empty]
\addplot[fill=cCascade, draw=none, error bars/.cd, y dir=both, y explicit] coordinates
  {(Probe,26.3) +- (0,14.5) (Surface,19.3) +- (0,11.7) (Stacking,26.8) +- (0,13.0)};
\addplot[fill=cSingle, draw=none, error bars/.cd, y dir=both, y explicit] coordinates
  {(Probe,14.6) +- (0,8.0) (Surface,5.2) +- (0,4.6) (Stacking,14.3) +- (0,7.9)};
\nextgroupplot[title={WebShop Qwen-2.5}, ymin=0, ymax=32,
  ytick={0,10,20,30}]
\addplot[fill=cCascade, draw=none, error bars/.cd, y dir=both, y explicit] coordinates
  {(Probe,18.7) +- (0,6.1) (Surface,10.3) +- (0,6.6) (Stacking,20.6) +- (0,8.2)};
\addplot[fill=cSingle, draw=none, error bars/.cd, y dir=both, y explicit] coordinates
  {(Probe,6.6) +- (0,3.6) (Surface,4.7) +- (0,4.9) (Stacking,8.0) +- (0,5.2)};
\nextgroupplot[title={WebShop Llama}, ymin=0, ymax=27,
  ytick={0,10,20}]
\addplot[fill=cCascade, draw=none, error bars/.cd, y dir=both, y explicit] coordinates
  {(Probe,11.4) +- (0,8.7) (Surface,7.6) +- (0,5.9) (Stacking,13.6) +- (0,9.6)};
\addplot[fill=cSingle, draw=none, error bars/.cd, y dir=both, y explicit] coordinates
  {(Probe,2.1) +- (0,2.6) (Surface,0.5) +- (0,0.7) (Stacking,1.1) +- (0,1.1)};
\nextgroupplot[title={WebShop Qwen3}, ymin=0, ymax=60,
  ytick={0,20,40,60}]
\addplot[fill=cCascade, draw=none, error bars/.cd, y dir=both, y explicit] coordinates
  {(Probe,41.6) +- (0,15.7) (Surface,9.9) +- (0,4.0) (Stacking,41.5) +- (0,15.8)};
\addplot[fill=cSingle, draw=none, error bars/.cd, y dir=both, y explicit] coordinates
  {(Probe,3.1) +- (0,1.4) (Surface,3.4) +- (0,3.8) (Stacking,3.8) +- (0,4.2)};
\end{groupplot}
\node[anchor=south] at ($(group c1r1.north)!0.5!(group c3r1.north)+(0,0.35cm)$)
  {\pgfplotslegendfromname{scorerlegend}};
\end{tikzpicture}
  \caption{Scorer ablation at target global recall $0.95$ across the
  full $2\times3$ matrix (mean $\pm$ one standard deviation over
  20 seeds). ``Stacking'' concatenates surface features onto the
  activation probe.}
  \label{fig:scorer}
\end{figure*}
}

\section{Results}

\subsection{Internal States Predict Failure Before Behavior Does}
\label{sec:auc}

Figure~\ref{fig:auc} plots the cross-fitted post-generation probe AUC
for predicting eventual failure among episodes alive at each round,
for every cell. In every cell the probe becomes
strongly informative within the first rounds, with timing that varies
across cells; this heterogeneity is itself informative. TextCraft
Qwen-2.5-7B and Qwen3-1.7B are already highly
separable in the first two rounds (AUC $0.86$ and $0.81$ at round 1),
while WebShop Llama-3.2-3B becomes
informative only after round 2. WebShop Qwen3-1.7B is the strongest
transfer case: its AUC rises from 0.591 at round 1 to 0.896 at round
3 and above 0.92 at later gates. These differences explain why the
savings frontier in Table~\ref{tab:main} differs across cells,
and they are precisely why a \emph{searched} allocation should beat
any fixed decision point: early gates protect more remaining compute,
but some cells require one or two rounds before the
hidden-state signal is reliable, and the budget search discovers this
per cell without manual tuning.
For WebShop Llama-3.2-3B, the format-constrained instruction reduces
invalid actions. The present experiment jointly reflects format
compliance and task comprehension; its near-chance early AUC establishes
the limited outcome predictiveness of the monitored state at this stage,
while leaving their individual contributions unresolved.

\subsection{Main Results: Cascade vs.\ Single Gate}
\label{sec:matrix}

Table~\ref{tab:main} compares the searched cascade with two baselines,
the best single gate and uniform allocation, over the complete
$2\times3$ matrix.
For comparability, all use the same activation-based stacking scorer,
Clopper--Pearson gate, grouped splits, 20 seeds, and margin
$\delta=0.02$; Section~\ref{sec:scorers} separately isolates the value
of the activation signal. Figure~\ref{fig:recall} plots achieved
recall; the complete savings frontier is in the supplementary
material. Three observations stand out.

First, \emph{the cascade dominates both allocation baselines in every
cell and at every target}: 24 of 24 configurations. At the 90\%
target its advantage over the best single gate ranges from
$1.5\times$ (TextCraft Qwen-2.5-7B, $60.2\%$ vs.\ $40.1\%$) to
$8.8\times$ (WebShop Llama-3.2-3B, $24.7\%$ vs.\ $2.8\%$). The gap is
largest exactly where the signal matures late
(Figure~\ref{fig:auc}): a single gate must commit to one round,
whereas the cascade defers budget to the rounds where each cell
becomes predictable. The uniform allocation, forced to be equally
strict at recall-expensive early gates and low-yield late ones,
collapses to near-zero savings, confirming that savings depend
critically on both the amount and placement of the recall budget.

Second, \emph{achieved recall tracks its target throughout}
(Figure~\ref{fig:recall}): across all 24 configurations the cascade's
mean test recall deviates from its target by at most $0.024$, always
within one seed-level standard deviation, and errs on the
conservative side in TextCraft. The largest deviations occur on
WebShop Qwen-2.5-7B at loose targets, where validation successes are
scarcest; they are well within seed noise, and the independent
post-selection certificate of Section~\ref{sec:cert} exists precisely
to convert this empirical tracking into a formal guarantee whenever
one is required. The savings in Table~\ref{tab:main} are thus attained
under honest, pre-specified risk control.

Third, \emph{savings follow signal quality more closely than success rate}.
The strongest cells (TextCraft Qwen-2.5-7B at $60.2\%$, WebShop
Qwen3-1.7B at $54.9\%$) are those whose probes are most separable in
the rounds that guard the most remaining compute; weaker cells retain
smaller but still positive savings while preserving their recall floor.

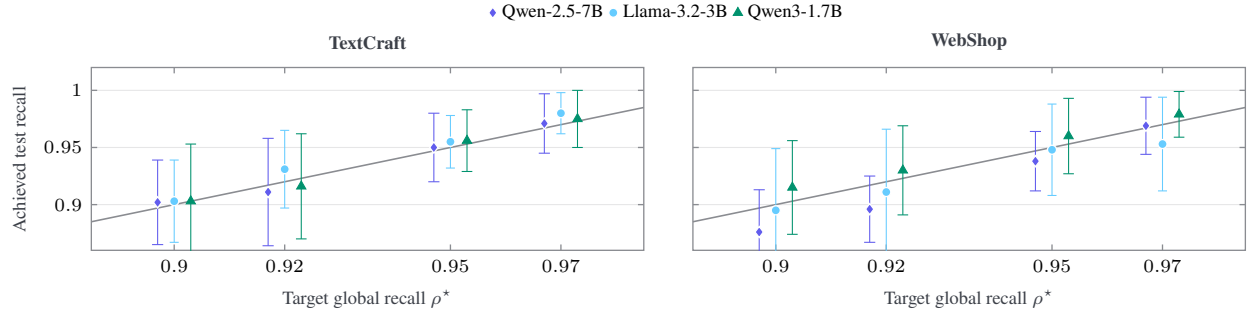
\begin{figure*}[t]
  \centering
  \begin{tikzpicture}
\begin{groupplot}[
  paper,
  group style={group size=2 by 1, horizontal sep=0.65cm,
    ylabels at=edge left, yticklabels at=edge left},
  width=0.50\textwidth, height=4.0cm,
  xlabel={Target global recall $\rho^\star$},
  ylabel={Achieved test recall},
  xmin=0.885, xmax=0.985, ymin=0.86, ymax=1.02,
  xtick={0.90,0.92,0.95,0.97},
  x tick label style={font=\scriptsize, /pgf/number format/.cd, fixed, precision=2},
  legend columns=3]
\nextgroupplot[title={TextCraft}, legend to name=recalllegend]
\addplot[cInk!60, line width=0.6pt, forget plot, domain=0.885:0.985, samples=2] {x};
\addplot[modelQ, only marks, error bars/.cd, y dir=both, y explicit] coordinates
  {(0.897,0.902) +- (0,0.037) (0.917,0.911) +- (0,0.047)
   (0.947,0.950) +- (0,0.030) (0.967,0.971) +- (0,0.026)};
\addplot[modelL, only marks, error bars/.cd, y dir=both, y explicit] coordinates
  {(0.90,0.903) +- (0,0.036) (0.92,0.931) +- (0,0.034)
   (0.95,0.955) +- (0,0.023) (0.97,0.980) +- (0,0.018)};
\addplot[cGreen, only marks, mark=triangle*, mark size=2pt,
  error bars/.cd, y dir=both, y explicit] coordinates
  {(0.903,0.903) +- (0,0.050) (0.923,0.916) +- (0,0.046)
   (0.953,0.956) +- (0,0.027) (0.973,0.975) +- (0,0.025)};
\legend{Qwen-2.5-7B, Llama-3.2-3B, Qwen3-1.7B}
\nextgroupplot[title={WebShop}]
\addplot[cInk!60, line width=0.6pt, forget plot, domain=0.885:0.985, samples=2] {x};
\addplot[modelQ, only marks, error bars/.cd, y dir=both, y explicit] coordinates
  {(0.897,0.876) +- (0,0.037) (0.917,0.896) +- (0,0.029)
   (0.947,0.938) +- (0,0.026) (0.967,0.969) +- (0,0.025)};
\addplot[modelL, only marks, error bars/.cd, y dir=both, y explicit] coordinates
  {(0.90,0.895) +- (0,0.054) (0.92,0.911) +- (0,0.055)
   (0.95,0.948) +- (0,0.040) (0.97,0.953) +- (0,0.041)};
\addplot[cGreen, only marks, mark=triangle*, mark size=2pt,
  error bars/.cd, y dir=both, y explicit] coordinates
  {(0.903,0.915) +- (0,0.041) (0.923,0.930) +- (0,0.039)
   (0.953,0.960) +- (0,0.033) (0.973,0.979) +- (0,0.020)};
\end{groupplot}
\node[anchor=south] at ($(group c1r1.north)!0.5!(group c2r1.north)+(0,0.35cm)$)
  {\pgfplotslegendfromname{recalllegend}};
\end{tikzpicture}
  \caption{Achieved global success recall of the searched cascade
  versus its target in the full $2\times3$ matrix (mean $\pm$ one
  standard deviation over 20 seeds; points are horizontally jittered
  for legibility). The diagonal marks exact targeting; every
  configuration lands within one standard deviation of its target.}
  \label{fig:recall}
\end{figure*}

\subsection{Does the Signal Require Internal Access?}
\label{sec:scorers}

Figure~\ref{fig:scorer} fixes the target at $0.95$ and swaps only the
scorer inside the otherwise-identical pipeline. Two conclusions
emerge. First, \emph{behavior-only monitoring is consistently weaker
than an activation probe}, most sharply on WebShop Qwen3-1.7B
($41.6\%$ vs.\ $9.9\%$ saved, a $4.2\times$ gap) and TextCraft
Qwen-2.5-7B ($42.2\%$ vs.\ $24.8\%$). Second, \emph{stacking surface
features onto the probe matches it and yields no further improvement}
in every
cell, indicating that whatever the surface scorer knows, the hidden
states already encode; the converse is clearly false. For the
behavioral features tested here, internal activations therefore retain
additional predictive information. A supplementary diagnostic at the strictest target
(technical appendix, ``Scorer Robustness at Strict Recall Targets'')
finds that the lower-dimensional probe additionally transfers more
reliably from validation to test when recall headroom is small, so we
recommend probe-only scoring when targets are strict. This ablation
compares against specified low-cost features; structured prefix
monitors remain an interesting extension within the same cascade.

\subsection{Stress Test: Failing Safe in a Low-Success Regime}
\label{sec:alfworld}

A reliable recall-controlled abort policy must also preserve successes
when the signal is weak. We use ALFWorld \citep{shridhar2021alfworld}
as a deliberate boundary test of this property. On 477 Qwen-2.5-7B
rollouts from 120 tasks, the early internal signal recurs in a third,
harder environment (probe AUC $0.815$), and at the nominal $0.95$
target the cascade saves $10.7\%$ versus $3.3\%$ for a single gate,
with mean recall $0.943$. For the weaker Qwen3-1.7B (AUC $0.764$),
successful calibration examples are scarce, and the cascade contracts
to $1$--$4\%$ savings. This is the designed behavior: when the data
cannot support aggressive aborting, the recall-calibrated gates
abstain and the policy approaches a no-op, prioritizing successes.
These experiments are recalibrated in-domain; zero-shot transfer
remains untested. Full counts and
per-target results appear in the supplementary material.

\subsection{The Cost of Independent Post-Selection Certification}
\label{sec:cert}

After margin-based search, a frozen cascade can be certified on an
independent sample ($\alpha_m=0.05$), and the certificate makes the
data cost of strict promises explicit. Figure~\ref{fig:cert} plots
the required number of successful episodes,
$n_{\mathrm{pos}}\ge\ln\alpha_m/\ln\rho^\star$. With 114 successes,
targets up to $\approx0.974$ are certifiable; $0.98$ and $0.99$ require
roughly 149 and 299, independent of scorer quality. When the sample is
insufficient, the valid choices are to abstain, collect more data, or
relax the target. This transparency makes the data requirement
operational: the same machinery that saves compute also reports,
before deployment, which recall promises the available data support
and prescribes exactly how much additional data a stricter promise
costs.

\relatedworksection

\section{Discussion and Limitations}

\subsubsection{Statistical control and selection.}
The default margin provides empirical control, validated in all 24
matrix configurations and by the supplementary margin sweep; an
independent post-selection sample upgrades a frozen cascade to a
formal distribution-free guarantee. Formal validity comes from the
independent certificate, while reused search data support empirical
selection. When successes are scarce, grouped
splits can select different discrete budgets; more successful tasks,
repeated grouped selection, and certification respectively reduce
selection variance and control the final policy's recall.

Our coarse grid fails conservatively: a suboptimal feasible point
preserves recall at the expense of savings. Finer search may improve a stepwise
frontier, although exhaustive cost grows as $|\mathcal{T}|^{R_g}$;
coordinate, greedy, or beam search are natural alternatives. We fixed
$\delta=0.02$ a priori, near the binomial standard error at our scale
($\rho^\star=0.95$, $n_{\mathrm{pos}}\approx114$). Under distribution
shift, the probe, thresholds, and global control must all be refreshed
on labeled data.

\subsubsection{Scope and system realization.}
Our balanced evidence spans two environments, three models, and 800
episodes per cell, with ALFWorld as a boundary test. Grouped splits
avoid task leakage but remain in-distribution; category-held-out tasks
are a natural next test. Distribution shift may also require
reselecting the pilot-fixed layer, and multi-layer features or later gates for longer
trajectories would require recalibration.

Teacher-forced replay supports our offline evaluation. For online use,
vLLM provides native selected-layer hidden-state extraction during
inference\footnote{\url{https://vllm.ai/blog/2026-03-30-extract-hidden-states}};
our monitor consumes one final-token vector per turn. Systems cost remains
nonzero because the documented extraction path persists activations to
shared storage and currently requires chunked prefill to be disabled. Our efficiency
metric is therefore generated-token savings; wall-clock latency and
dollar cost remain unmeasured.

\subsubsection{What aborted compute buys.}
The present evaluation measures saved compute. Reallocating it to
retries would create a test-time-scaling policy whose reward and
systems effects require end-to-end measurement.

\section{Conclusion}

Across a $2\times3$ matrix, internal activations predict eventual agent
failure before behavioral features become comparably informative. A
recall-calibrated cascade converts this signal into up to $60.2\%$
token savings at 90\% global recall, beats the best single gate in all
24 configurations, and approaches no-op when evidence weakens.
Independent certification then identifies the recall promises supported
by the available data. This yields an auditable
route to adaptive agent inference without silently sacrificing task
success.

\bibliography{references}

@book{vovk2005algorithmic,
  author={Vovk, Vladimir and Gammerman, Alexander and Shafer, Glenn},
  title={Algorithmic Learning in a Random World},
  publisher={Springer},
  year={2005}
}

@inproceedings{papadopoulos2002inductive,
  author={Papadopoulos, Harris and Proedrou, Kostas and Vovk, Volodya and Gammerman, Alex},
  title={Inductive Confidence Machines for Regression},
  booktitle={European Conference on Machine Learning},
  year={2002}
}

@article{lei2018distribution,
  author={Lei, Jing and G'Sell, Max and Rinaldo, Alessandro and Tibshirani, Ryan J and Wasserman, Larry},
  title={Distribution-Free Predictive Inference for Regression},
  journal={Journal of the American Statistical Association},
  volume={113}, number={523}, pages={1094--1111},
  year={2018}
}

@article{angelopoulos2023gentle,
  author={Angelopoulos, Anastasios N and Bates, Stephen},
  title={Conformal Prediction: A Gentle Introduction},
  journal={Foundations and Trends in Machine Learning},
  volume={16}, number={4}, pages={494--591},
  year={2023}
}

@article{bates2021distribution,
  author={Bates, Stephen and Angelopoulos, Anastasios and Lei, Lihua and Malik, Jitendra and Jordan, Michael I},
  title={Distribution-Free, Risk-Controlling Prediction Sets},
  journal={Journal of the ACM},
  volume={68}, number={6}, pages={1--34},
  year={2021}
}

@article{angelopoulos2025learn,
  author={Angelopoulos, Anastasios N and Bates, Stephen and Cand{\`e}s, Emmanuel J and Jordan, Michael I and Lei, Lihua},
  title={Learn then Test: Calibrating Predictive Algorithms to Achieve Risk Control},
  journal={The Annals of Applied Statistics},
  volume={19}, number={2}, pages={1641--1662},
  year={2025}
}

@inproceedings{angelopoulos2024conformal,
  author={Angelopoulos, Anastasios N and Bates, Stephen and Fisch, Adam and Lei, Lihua and Schuster, Tal},
  title={Conformal Risk Control},
  booktitle={International Conference on Learning Representations},
  year={2024}
}

@inproceedings{quach2024conformal,
  author={Quach, Victor and Fisch, Adam and Schuster, Tal and Yala, Adam and Sohn, Jae Ho and Jaakkola, Tommi S and Barzilay, Regina},
  title={Conformal Language Modeling},
  booktitle={International Conference on Learning Representations},
  year={2024}
}

@inproceedings{mohri2024language,
  author={Mohri, Christopher and Hashimoto, Tatsunori},
  title={Language Models with Conformal Factuality Guarantees},
  booktitle={International Conference on Machine Learning},
  year={2024}
}

@inproceedings{cherian2024large,
  author={Cherian, John J and Gibbs, Isaac and Cand{\`e}s, Emmanuel J},
  title={Large Language Model Validity via Enhanced Conformal Prediction Methods},
  booktitle={Advances in Neural Information Processing Systems},
  year={2024}
}

@inproceedings{ren2023robots,
  author={Ren, Allen Z and Dixit, Anushri and Bodrova, Alexandra and Singh, Sumeet and Tu, Stephen and Brown, Noah and Xu, Peng and Takayama, Leila and Xia, Fei and Varley, Jake and Xu, Zhenjia and Sadigh, Dorsa and Zeng, Andy and Majumdar, Anirudha},
  title={Robots That Ask for Help: Uncertainty Alignment for Large Language Model Planners},
  booktitle={Conference on Robot Learning},
  year={2023}
}

@inproceedings{schuster2022confident,
  author={Schuster, Tal and Fisch, Adam and Gupta, Jai and Dehghani, Mostafa and Bahri, Dara and Tran, Vinh Q and Tay, Yi and Metzler, Donald},
  title={Confident Adaptive Language Modeling},
  booktitle={Advances in Neural Information Processing Systems},
  year={2022}
}

@article{alain2016understanding,
  author={Alain, Guillaume and Bengio, Yoshua},
  title={Understanding Intermediate Layers Using Linear Classifier Probes},
  journal={arXiv preprint arXiv:1610.01644},
  year={2016}
}

@article{belinkov2022probing,
  author={Belinkov, Yonatan},
  title={Probing Classifiers: Promises, Shortcomings, and Advances},
  journal={Computational Linguistics},
  volume={48}, number={1}, pages={207--219},
  year={2022}
}

@article{kadavath2022language,
  author={Kadavath, Saurav and Conerly, Tom and Askell, Amanda and Henighan, Tom and Drain, Dawn and others},
  title={Language Models (Mostly) Know What They Know},
  journal={arXiv preprint arXiv:2207.05221},
  year={2022}
}

@inproceedings{azaria2023internal,
  author={Azaria, Amos and Mitchell, Tom},
  title={The Internal State of an {LLM} Knows When It's Lying},
  booktitle={Findings of the Association for Computational Linguistics: EMNLP},
  year={2023}
}

@inproceedings{burns2023discovering,
  author={Burns, Collin and Ye, Haotian and Klein, Dan and Steinhardt, Jacob},
  title={Discovering Latent Knowledge in Language Models Without Supervision},
  booktitle={International Conference on Learning Representations},
  year={2023}
}

@inproceedings{marks2023geometry,
  author={Marks, Samuel and Tegmark, Max},
  title={The Geometry of Truth: Emergent Linear Structure in Large Language Model Representations of True/False Datasets},
  booktitle={Conference on Language Modeling},
  year={2024}
}

@inproceedings{orgad2025llms,
  author={Orgad, Hadas and Toker, Michael and Gekhman, Zorik and Reichart, Roi and Szpektor, Idan and Kotek, Hadas and Belinkov, Yonatan},
  title={{LLM}s Know More Than They Show: On the Intrinsic Representation of {LLM} Hallucinations},
  booktitle={International Conference on Learning Representations},
  year={2025}
}

@article{kossen2024semantic,
  author={Kossen, Jannik and Han, Jiatong and Razzak, Muhammed and Schut, Lisa and Malik, Shreshth and Gal, Yarin},
  title={Semantic Entropy Probes: Robust and Cheap Hallucination Detection in {LLM}s},
  journal={arXiv preprint arXiv:2406.15927},
  year={2024}
}

@article{zhang2025reasoning,
  author={Zhang, Anqi and Chen, Yulin and Pan, Jane and Zhao, Chen and Panda, Aurojit and Li, Jinyang and He, He},
  title={Reasoning Models Know When They're Right: Probing Hidden States for Self-Verification},
  journal={arXiv preprint arXiv:2504.05419},
  year={2025}
}

@inproceedings{afzal2025knowing,
  author={Afzal, Anum and Matthes, Florian and Chechik, Gal and Ziser, Yftah},
  title={Knowing Before Saying: {LLM} Representations Encode Information About Chain-of-Thought Success Before Completion},
  booktitle={Findings of the Association for Computational Linguistics: ACL},
  year={2025}
}

@inproceedings{jiang2025runaway,
  author={Lu, Qingyu and Ding, Liang and Cao, Siyi and Liu, Xuebo and Zhang, Kanjian and Zhang, Jinxia and Tao, Dacheng},
  title={Runaway is Ashamed, But Helpful: On the Early-Exit Behavior of Large Language Model-based Agents in Embodied Environments},
  booktitle={Findings of the Association for Computational Linguistics: EMNLP},
  year={2025}
}

@article{deng2025agentdebug,
  author={Zhu, Kunlun and Liu, Zijia and Li, Bingxuan and Tian, Muxin and Yang, Yingxuan and Zhang, Jiaxun and Han, Pengrui and Xie, Qipeng and Cui, Fuyang and Zhang, Weijia and Ma, Xiaoteng and Yu, Xiaodong and Ramesh, Gowtham and Wu, Jialian and Liu, Zicheng and Lu, Pan and Zou, James and You, Jiaxuan},
  title={Where {LLM} Agents Fail and How They Can Learn from Failures},
  journal={arXiv preprint arXiv:2509.25370},
  year={2025}
}

@article{zhang2026foresight,
  author={Zhang, Boxuan and Zhu, Jianing and Shi, Zeru and Liu, Dongfang and Tang, Ruixiang},
  title={AgentForesight: Online Auditing for Early Failure Prediction in Multi-Agent Systems},
  journal={arXiv preprint arXiv:2605.08715},
  year={2026}
}

@article{anon2026alphastop,
  author={Baidya, Avinash and Liang, Xinran and Guo, Ruocheng and Gao, Xiang and Das, Kamalika},
  title={When Evidence is Sparse: Weakly Supervised Early Failure Alerting in Dialogs and {LLM}-Agent Trajectories},
  journal={arXiv preprint arXiv:2606.05414},
  year={2026}
}

@article{anon2026bagen,
  author={Lin, Yuxiang and Wang, Zihan and Liu, Mengyang and Shan, Yuxuan and Bai, Longju and Zhang, Junyao and Jin, Xing and Chen, Boshan and Su, Jinyan and Wang, Xingyao and Pei, Jiaxin and Li, Manling},
  title={{BAGEN}: Are {LLM} Agents Budget-Aware?},
  journal={arXiv preprint arXiv:2606.00198},
  year={2026},
  url={https://arxiv.org/abs/2606.00198}
}

@article{mittapalli2026trace,
  author={Mittapalli, Vijitha and Dani, Shreyaa Jayant and Pilli, Satya Srujana and Ansu, Snigdha and Teymoorianfard, Mohammadreza and Dernoncourt, Franck and Chen, Hongjie and Wang, Yu and Rossi, Ryan A. and Ahmed, Nesreen K.},
  title={{TRACE}: Trajectory Reasoning through Adaptive Cross-Step Evidence Aggregation for {LLM} Agents},
  journal={arXiv preprint arXiv:2606.07054},
  year={2026},
  url={https://arxiv.org/abs/2606.07054}
}

@article{dellibarda2026poirot,
  author={Dellibarda Varela, I{\~n}aki and Sendra-Arranz, R. and Romero-Sorozabal, Pablo and Valverde-Garc{\'i}a, J. M. and Laudanski, Annemarie F. and Guti{\'e}rrez, {\'A}lvaro and Rocon, Eduardo and Cebrian, Manuel},
  title={{POIROT}: Interrogating Agents for Failure Detection in Multi-Agent Systems},
  journal={arXiv preprint arXiv:2606.02282},
  year={2026},
  url={https://arxiv.org/abs/2606.02282}
}

@inproceedings{romer2025fiper,
  author={R{\"o}mer, Ralf and Kobras, Adrian and Worbis, Luca and Schoellig, Angela P.},
  title={Failure Prediction at Runtime for Generative Robot Policies},
  booktitle={Advances in Neural Information Processing Systems},
  year={2025},
  url={https://arxiv.org/abs/2510.09459}
}

@article{anon2026observability,
  author={Li, Xianyou and Yan, Weiran and Wu, Yichao and Liang, Penghao and Yuan, Mengwei and Liu, Jianan and Yang, Jing},
  title={Early Diagnosis of Wasted Computation in Multi-Agent {LLM} Systems via Failure-Aware Observability},
  journal={arXiv preprint arXiv:2606.01365},
  year={2026}
}

@article{anon2026conformal_interp,
  author={Padhi, Trilok and Kaur, Ramneet and Agarwal, Krishiv and Cobb, Adam D. and Elenius, Daniel and Acharya, Manoj and Samplawski, Colin and Berenbeim, Alexander M. and Bastian, Nathaniel D. and Jha, Susmit and Roy, Anirban},
  title={From Actions to Understanding: Conformal Interpretability of Temporal Concepts in {LLM} Agents},
  journal={arXiv preprint arXiv:2604.19775},
  year={2026}
}

@inproceedings{ji2024llm,
  author={Ji, Ziwei and Chen, Delong and Ishii, Etsuko and Cahyawijaya, Samuel and Bang, Yejin and Wilie, Bryan and Fung, Pascale},
  title={{LLM} Internal States Reveal Hallucination Risk Faced With a Query},
  booktitle={Proceedings of the 7th BlackboxNLP Workshop: Analyzing and Interpreting Neural Networks for NLP},
  year={2024}
}

@inproceedings{aggarwal2023lets,
  author={Aggarwal, Pranjal and Madaan, Aman and Yang, Yiming and Mausam},
  title={Let's Sample Step by Step: Adaptive-Consistency for Efficient Reasoning and Coding with {LLM}s},
  booktitle={Empirical Methods in Natural Language Processing},
  year={2023}
}

@inproceedings{li2024escape,
  author={Li, Yiwei and Yuan, Peiwen and Feng, Shaoxiong and Pan, Boyuan and Wang, Xinglin and Sun, Bin and Wang, Heda and Li, Kan},
  title={Escape Sky-High Cost: Early-Stopping Self-Consistency for Multi-Step Reasoning},
  booktitle={International Conference on Learning Representations},
  year={2024}
}

@article{manvi2024adaptive,
  author={Manvi, Rohin and Singh, Anikait and Ermon, Stefano},
  title={Adaptive Inference-Time Compute: {LLM}s Can Predict If They Can Do Better, Even Mid-Generation},
  journal={arXiv preprint arXiv:2410.02725},
  year={2024}
}

@article{chen2023frugalgpt,
  author={Chen, Lingjiao and Zaharia, Matei and Zou, James},
  title={Frugal{GPT}: How to Use Large Language Models While Reducing Cost and Improving Performance},
  journal={arXiv preprint arXiv:2305.05176},
  year={2023}
}

@article{zellinger2025cost,
  author={Zellinger, Michael J. and Liu, Rex and Thomson, Matt},
  title={Cost-Saving {LLM} Cascades with Early Abstention},
  journal={arXiv preprint arXiv:2502.09054},
  year={2025}
}

@inproceedings{lightman2024lets,
  author={Lightman, Hunter and Kosaraju, Vineet and Burda, Yura and Edwards, Harri and Baker, Bowen and Lee, Teddy and Leike, Jan and Schulman, John and Sutskever, Ilya and Cobbe, Karl},
  title={Let's Verify Step by Step},
  booktitle={International Conference on Learning Representations},
  year={2024}
}

@inproceedings{zhang2024rest,
  author={Zhang, Dan and Zhoubian, Sining and Hu, Ziniu and Yue, Yisong and Dong, Yuxiao and Tang, Jie},
  title={Re{ST}-{MCTS}*: {LLM} Self-Training via Process Reward Guided Tree Search},
  booktitle={Advances in Neural Information Processing Systems},
  year={2024}
}

@inproceedings{xia2025agentrm,
  author={Xia, Yu and Fan, Jingru and Chen, Weize and Yan, Siyu and Cong, Xin and Zhang, Zhong and Lu, Yaxi and Lin, Yankai and Liu, Zhiyuan and Sun, Maosong},
  title={Agent{RM}: Enhancing Agent Generalization with Reward Modeling},
  booktitle={Proceedings of the 63rd Annual Meeting of the Association for Computational Linguistics (Volume 1: Long Papers)},
  year={2025}
}

@article{anon2025refrain,
  author={Sun, Renliang and Cheng, Wei and Li, Dawei and Chen, Haifeng and Wang, Wei},
  title={Stop When Enough: Adaptive Early-Stopping for Chain-of-Thought Reasoning},
  journal={arXiv preprint arXiv:2510.10103},
  year={2025}
}

@article{mehta2026premature,
  title={When Agents Commit Too Soon: Diagnosing Premature Commitment in {LLM} Agents},
  author={Mehta, Aman},
  journal={arXiv preprint arXiv:2606.22936},
  year={2026}
}

@article{kim2026atropos,
  title={Atropos: Improving Cost-Benefit Trade-off of {LLM}-based Agents under Self-Consistency with Early Termination and Model Hotswap},
  author={Kim, Naryeong and Yoo, Shin},
  journal={arXiv preprint arXiv:2604.15075},
  year={2026}
}

@article{davidov2026quit,
  title={Knowing When to Quit: A Principled Framework for Dynamic Abstention in {LLM} Reasoning},
  author={Davidov, Hen and Cohen, Nachshon and Kalinsky, Oren and Fairstein, Yaron and Kushilevitz, Guy and Yazdi, Ram and Rebeschini, Patrick},
  journal={arXiv preprint arXiv:2604.18419},
  year={2026}
}

@inproceedings{shen2025masc,
  title={Metacognitive Self-Correction for Multi-Agent System via Prototype-Guided Next-Execution Reconstruction},
  author={Shen, Xu and Zhang, Qi and Wang, Song and Tan, Zhen and Zhao, Xinyu and Yao, Laura and Tadiparthi, Vaishnav and Mahjoub, Hossein Nourkhiz and {Moradi Pari}, Ehsan and Lee, Kwonjoon and Chen, Tianlong},
  booktitle={Findings of the Association for Computational Linguistics: ACL 2026},
  year={2026}
}

@article{barke2026agentrx,
  title={{AgentRx}: Diagnosing {AI} Agent Failures from Execution Trajectories},
  author={Barke, Shraddha and Goyal, Arnav and Khare, Alind and Singh, Avaljot and Nath, Suman and Bansal, Chetan},
  journal={arXiv preprint arXiv:2602.02475},
  year={2026}
}

@article{chen2026signals,
  title={Signals: Trajectory Sampling and Triage for Agentic Interactions},
  author={Chen, Shuguang and others},
  journal={arXiv preprint arXiv:2604.00356},
  year={2026}
}

@article{mao2025escot,
  title={Early Stopping Chain-of-Thoughts in Large Language Models},
  author={Mao, Minjia and Yin, Bowen and Zhu, Yu and Fang, Xiao},
  journal={arXiv preprint arXiv:2509.14004},
  year={2025}
}

@article{chen2023eellm,
  title={{EE-LLM}: Large-Scale Training and Inference of Early-Exit Large Language Models with {3D} Parallelism},
  author={Chen, Yanxi and others},
  journal={arXiv preprint arXiv:2312.04916},
  year={2023}
}

@article{miao2024efficient,
  title={An Efficient Inference Framework for Early-Exit Large Language Models},
  author={Miao, Ruijie and Yan, Yihan and Yao, Xinshuo and Yang, Tong},
  journal={arXiv preprint arXiv:2407.20272},
  year={2024}
}

@misc{meta2024llama32,
  author={{Meta}},
  title={{Llama-3.2-3B} Model Card},
  year={2024},
  month={September},
  howpublished={\url{https://huggingface.co/meta-llama/Llama-3.2-3B}}
}

@article{qwen2025qwen25,
  author={{Qwen Team}},
  title={{Qwen2.5} Technical Report},
  journal={arXiv preprint arXiv:2412.15115},
  year={2025}
}

@inproceedings{prasad2024adapt,
  author={Prasad, Archiki and Koller, Alexander and Hartmann, Mareike and Clark, Peter and Sabharwal, Ashish and Bansal, Mohit and Khot, Tushar},
  title={{ADaPT}: As-Needed Decomposition and Planning with Language Models},
  booktitle={Findings of the Association for Computational Linguistics: NAACL 2024},
  year={2024}
}

@inproceedings{shridhar2021alfworld,
  author={Shridhar, Mohit and Yuan, Xingdi and C{\^o}t{\'e}, Marc-Alexandre and Bisk, Yonatan and Trischler, Adam and Hausknecht, Matthew},
  title={{ALFWorld}: Aligning Text and Embodied Environments for Interactive Learning},
  booktitle={International Conference on Learning Representations},
  year={2021},
  url={https://openreview.net/forum?id=0IOX0YcCdTn}
}

@article{shao2024deepseekmath,
  author={Shao, Zhihong and Wang, Peiyi and Zhu, Qihao and Xu, Runxin and Song, Junxiao and Bi, Xiao and Zhang, Haowei and Zhang, Mingchuan and Li, Y. K. and Wu, Y. and Guo, Daya},
  title={{DeepSeekMath}: Pushing the Limits of Mathematical Reasoning in Open Language Models},
  journal={arXiv preprint arXiv:2402.03300},
  year={2024}
}

@article{deepseekai2025r1,
  author={{DeepSeek-AI}},
  title={{DeepSeek-R1}: Incentivizing Reasoning Capability in {LLM}s via Reinforcement Learning},
  journal={Nature},
  volume={645},
  pages={633--638},
  year={2025}
}

@article{xi2024agentgym,
  author={Xi, Zhiheng and Ding, Yiwen and Chen, Wenxiang and Hong, Boyang and Guo, Honglin and Wang, Junzhe and Yang, Dingwen and Liao, Chenyang and Guo, Xin and He, Wei and Gao, Songyang and Chen, Lu and Zheng, Rui and Zou, Yicheng and Gui, Tao and Zhang, Qi and Qiu, Xipeng and Huang, Xuanjing and Wu, Zuxuan and Jiang, Yu-Gang},
  title={{AgentGym}: Evolving Large Language Model-based Agents across Diverse Environments},
  journal={arXiv preprint arXiv:2406.04151},
  year={2024}
}

@inproceedings{guan2026futile,
  author={Guan, Xinyan and Zeng, Jiali and Xin, Chunlei and Lu, Yaojie and Lin, Hongyu and Han, Xianpei and Sun, Le and Meng, Fandong},
  title={Knowing When to Quit: Diagnosing and Training {LLM}s to Abort Futile Reasoning},
  booktitle={Findings of the Association for Computational Linguistics: ACL 2026},
  pages={16823--16835},
  year={2026},
  doi={10.18653/v1/2026.findings-acl.830},
  url={https://aclanthology.org/2026.findings-acl.830/}
}

@article{shrivastava2026semantic,
  author={Shrivastava, Sahil},
  title={Semantic Early-Stopping for Iterative {LLM} Agent Loops},
  journal={arXiv preprint arXiv:2606.27009},
  year={2026}
}

@article{ma2026memprobe,
  author={Ma, Enze and Zhou, Yufan and Huang, Wei-Chieh and Yang, Jie and Ma, Huanhuan and Wang, Zixuan and Li, Chengze and Miao, Chunyu and Yu, Philip S. and Wang, Zhen},
  title={{MEMPROBE}: Probing Long-Term Agent Memory via Hidden User-State Recovery},
  journal={arXiv preprint arXiv:2606.24595},
  year={2026}
}

@article{pham2026agentstop,
  author={Pham, Dzung and Katevas, Kleomenis and Shamsabadi, Ali Shahin and Haddadi, Hamed},
  title={{AgentStop}: Terminating Local {AI} Agents Early to Save Energy in Consumer Devices},
  journal={arXiv preprint arXiv:2605.15206},
  year={2026},
  url={https://arxiv.org/abs/2605.15206}
}

@article{zhao2026grade,
  author={Zhao, Yue},
  title={{GRADE}: Graph Representation of {LLM} Agent Dependency and Execution},
  journal={arXiv preprint arXiv:2606.22741},
  year={2026},
  url={https://arxiv.org/abs/2606.22741}
}

@article{huang2026prefixguard,
  author={Huang, Xinmiao and Hu, Jinwei and Roy, Rajarshi and Wu, Changshun and Dong, Yi and Huang, Xiaowei},
  title={{PrefixGuard}: From {LLM}-Agent Traces to Online Failure-Warning Monitors},
  journal={arXiv preprint arXiv:2605.06455},
  year={2026},
  url={https://arxiv.org/abs/2605.06455}
}

@article{qu2024recursive,
  author={Qu, Yuxiao and Zhang, Tianjun and Garg, Naman and Kumar, Aviral},
  title={Recursive Introspection: Teaching Language Model Agents How to Self-Improve},
  journal={arXiv preprint arXiv:2407.18219},
  year={2024},
  url={https://arxiv.org/abs/2407.18219}
}

\clearpage
\appendix

\twocolumn[
\begin{center}
{\Large\bfseries Appendix}
\end{center}
\vspace{0.75em}
]

\subsection*{A. Per-Layer Probe AUC Sweeps}
\label{app:layers}

Probe layers were fixed once per model and then used unchanged in all
recall-target and cascade evaluations. For Qwen-2.5-7B, a per-layer
sweep over layers $\{0, 2, \dots, 28\}$ was run on an independent
pilot set collected before the main experiments, selecting layer 20.
For Llama-3.2-3B, the sweep covered layers
$\{6, 10, 14, 18, 22, 26\}$ and selected layer 14. For Qwen3-1.7B,
the full-matrix run swept layers $\{4,8,\dots,28\}$ and selected
layer 28.

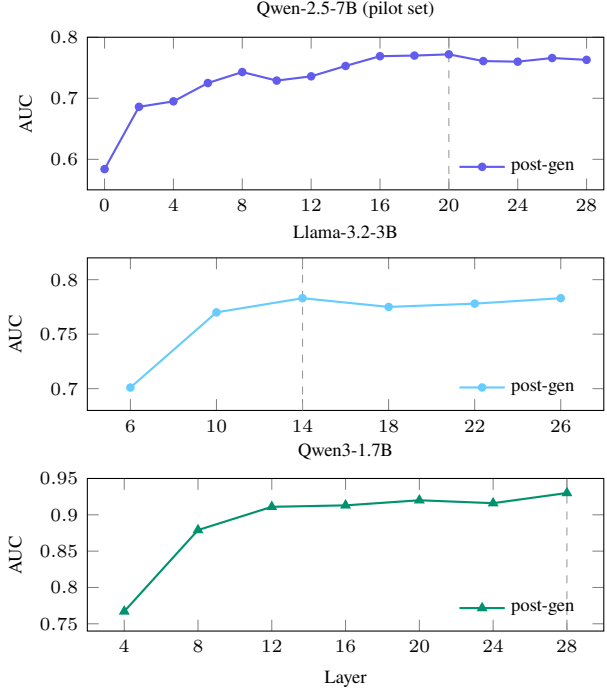
\begin{figure}[H]
  \centering
  \begin{tikzpicture}
\begin{groupplot}[
  group style={group size=1 by 3, vertical sep=0.9cm},
  width=\linewidth, height=3.6cm,
  ylabel={AUC},
  tick label style={font=\scriptsize},
  label style={font=\scriptsize},
  title style={font=\scriptsize, yshift=-0.5ex},
  legend style={font=\scriptsize, draw=none, fill=none},
  legend pos=south east]
\nextgroupplot[title={Qwen-2.5-7B (pilot set)}, xmin=-1, xmax=29,
  xtick={0,4,8,12,16,20,24,28}, ymin=0.55, ymax=0.80]
\addplot[cQwen, thick, mark=*, mark size=1.2pt] coordinates
  {(0,0.584)(2,0.686)(4,0.695)(6,0.725)(8,0.743)(10,0.729)(12,0.736)
   (14,0.753)(16,0.769)(18,0.770)(20,0.772)(22,0.761)(24,0.760)
   (26,0.766)(28,0.763)};
\addplot[cVanilla, dashed] coordinates {(20,0.55)(20,0.80)};
\legend{post-gen}
\nextgroupplot[title={Llama-3.2-3B}, xmin=4, xmax=28,
  xtick={6,10,14,18,22,26}, ymin=0.68, ymax=0.82]
\addplot[cLlama, thick, mark=*, mark size=1.2pt] coordinates
  {(6,0.701)(10,0.770)(14,0.783)(18,0.775)(22,0.778)(26,0.783)};
\addplot[cVanilla, dashed] coordinates {(14,0.68)(14,0.82)};
\legend{post-gen}
\nextgroupplot[title={Qwen3-1.7B}, xmin=2, xmax=30,
  xtick={4,8,12,16,20,24,28}, ymin=0.74, ymax=0.95,
  xlabel={Layer}]
\addplot[cGreen, thick, mark=triangle*, mark size=1.6pt] coordinates
  {(4,0.767)(8,0.879)(12,0.911)(16,0.913)(20,0.920)(24,0.916)(28,0.930)};
\addplot[cVanilla, dashed] coordinates {(28,0.74)(28,0.95)};
\legend{post-gen}
\end{groupplot}
\end{tikzpicture}
  \caption{Per-layer probe-AUC sweeps used to fix the probe layer for
  each agent. Qwen-2.5-7B: independent pilot set, layer 20.
  Llama-3.2-3B: layer 14. Qwen3-1.7B: layer 28.}
  \label{fig:layersweep}
\end{figure}

\newpage
\subsection*{B. Conformal-Quantile Gates}
\label{app:quantile}

Table~\ref{tab:quantile} compares the Clopper--Pearson gate of the
main text against the conformal-quantile variant, for a single
probe-based gate at round 1 with per-round target $0.95$ (20 seeds).
The quantile gate controls recall only on average over the calibration
draw, so with the small per-round calibration
sets available here its realized recall fluctuates below target,
violating it for Llama ($0.933$ vs.\ $0.95$) and Qwen3-1.7B
($0.940$ vs.\ $0.95$) while saving more compute. The
Clopper--Pearson gate is conservative (recall $0.966$--$0.977$ at
target $0.95$) but never violates in these round-1 comparisons. This is the
trade the main text accepts: the cascade's budget search recovers
much of the conservatism by spending recall where it is cheap,
while retaining a high-confidence per-gate guarantee.

\begin{table}[H]
\centering
\small
\setlength{\tabcolsep}{3pt}
\begin{tabular}{@{}llcc@{}}
\toprule
Model & Calibration & Recall & Saved (\%) \\
\midrule
Llama-3.2-3B & Clopper--Pearson & $0.972 \pm 0.013$ & $10.4 \pm 3.9$ \\
             & Quantile         & $0.933 \pm 0.029$ & $20.7 \pm 5.6$ \\
\midrule
Qwen-2.5-7B & Clopper--Pearson & $0.977 \pm 0.011$ & $17.4 \pm 5.5$ \\
            & Quantile         & $0.954 \pm 0.020$ & $24.7 \pm 4.8$ \\
\midrule
Qwen3-1.7B & Clopper--Pearson & $0.966 \pm 0.020$ & $14.6 \pm 8.0$ \\
           & Quantile         & $0.940 \pm 0.032$ & $27.7 \pm 9.8$ \\
\bottomrule
\end{tabular}
\caption{Single gate at round 1, per-round recall target $0.95$,
probe scorer, 20 seeds. Quantile calibration saves more but violates
the target on Llama-3.2-3B and Qwen3-1.7B; Clopper--Pearson is conservative
and meets it in all three model families.}
\label{tab:quantile}
\end{table}

\newpage
\subsection*{C. Scorer Robustness at Strict Recall Targets}
\label{app:strict}

At $\rho^\star = 0.95$, stacking surface features onto the probe
neither helps nor hurts. At the strictest target the comparison
changes character: with
$\rho^\star = 0.97$ the headroom between the target and perfect
recall is only $0.03$, shifting the focus to whether the budget selected
on validation still controls recall on test. Table~\ref{tab:strict} compares three scorers on
TextCraft with Qwen-2.5-7B at $\rho^\star = 0.97$: a logistic
activation probe, an MLP variant of the probe (identical features
with a nonlinear head: a single hidden layer of $256$ units on the
same standardized features), and the stacking scorer.

In mean savings the three scorers are statistically indistinguishable
at this target ($17.2\%$, $16.7\%$, and $15.5\%$, each with a
standard deviation above ten points; the largest paired difference,
MLP vs.\ stacking, is $1.7$ points with standard error $2.6$,
$t(19) = 0.67$), so the comparison rests on per-seed recall control.
There the scorers order consistently, though the differences are not
individually significant at $20$ seeds: the MLP probe meets the floor
in all $20$ seeds (minimum test recall $0.971$), the logistic probe
in $17$ of $20$ (minimum $0.960$), and stacking in $16$ of $20$
(minimum $0.957$). Because all three scorers are evaluated on the
same $20$ splits, the appropriate test is paired: in the extreme
comparison (MLP, $0$ violations, vs.\ stacking, $4$), all four
discordant seeds favor the MLP, and an exact one-sided sign test
gives $p = 0.062$, directional but not significant at this sample
size. Two of the violating seeds are shared by the logistic probe
and stacking, so split difficulty rather than scorer identity drives
most violations. (This batch shares its seed stream with the margin
sweep in Appendix~\ref{app:margin}; under the shift identity, the logistic row here
coincides seed-for-seed with that sweep's $\delta = 0.02$ entry at
this target, $3$ of $20$ in Table~\ref{tab:deltaviol}. The seed-for-seed
agreement provides a deterministic consistency check on the
implementation.) What distinguishes stacking is the
depth of its worst failure: its worst seed selects a budget whose
validation recall is $1.000$ yet achieves only $0.957$ on test, a
transfer gap of $0.043$ that exceeds the entire $0.03$ headroom
between the target and perfect recall, against a worst logistic gap
of $0.031$ and no violation at all for the MLP probe. Every violating
seed of either scorer selected a budget with validation recall at
least $0.991$, so these failures are invisible at search time; they
are precisely the selection optimism the margin is sized to absorb
in the margin-size sweep of Appendix~\ref{app:margin}. The
higher-dimensional stacked features give the $46{,}656$-candidate
budget search more room to overfit the validation split, and since
the surface features add no discriminative signal beyond the
activations, the extra dimensions buy variance without information.
The practical prescription is simple: when recall headroom is small,
prefer the lower-dimensional probe (or its MLP variant), which
delivers the same savings with tighter recall control.

This robustness batch uses the seed stream shared with the margin-size
sweep and serves as a separate diagnostic of strict-target behavior.

\begin{table}[H]
\centering
\resizebox{\linewidth}{!}{%
\begin{tabular}{@{}lcccc@{}}
\toprule
Scorer & Recall & Saved (\%) & Min recall & Seeds $\ge \rho^\star$ \\
\midrule
MLP probe      & $0.989 \pm 0.010$ & $17.2 \pm 10.8$ & $0.971$ & $20/20$ \\
Logistic probe & $0.985 \pm 0.011$ & $16.7 \pm 10.5$ & $0.960$ & $17/20$ \\
Stacking       & $0.986 \pm 0.014$ & $15.5 \pm 14.0$ & $0.957$ & $16/20$ \\
\bottomrule
\end{tabular}
}
\caption{Cascade at $\rho^\star = 0.97$ on TextCraft with
Qwen-2.5-7B ($20$ seeds,
test split; margin $\delta = 0.02$ as in the main experiments).
``Seeds $\ge \rho^\star$'' counts seeds whose test recall meets the
floor.}
\label{tab:strict}
\end{table}

\newpage
\subsection*{D. Margin Size Sweep}
\label{app:margin}

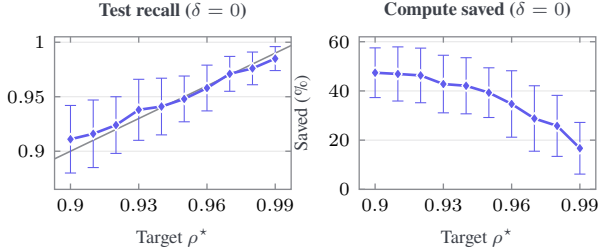
\begin{figure}[H]
  \centering
  \begin{tikzpicture}
\begin{groupplot}[
  paper,
  group style={group size=2 by 1, horizontal sep=0.9cm},
  width=0.56\linewidth, height=3.6cm,
  xmin=0.893, xmax=0.997, xtick={0.90,0.93,0.96,0.99},
  x tick label style={font=\scriptsize, /pgf/number format/.cd, fixed, precision=2},
  xlabel={Target $\rho^\star$}]
\nextgroupplot[title={Test recall ($\delta = 0$)},
  ymin=0.865, ymax=1.005,
  y tick label style={font=\scriptsize, /pgf/number format/.cd, fixed, precision=2}]
\addplot[cInk!60, line width=0.6pt, forget plot, domain=0.893:0.997, samples=2] {x};
\addplot[modelQ, error bars/.cd, y dir=both, y explicit] coordinates
  {(0.90,0.911) +- (0,0.031)
   (0.91,0.916) +- (0,0.031)
   (0.92,0.924) +- (0,0.026)
   (0.93,0.938) +- (0,0.028)
   (0.94,0.941) +- (0,0.026)
   (0.95,0.948) +- (0,0.021)
   (0.96,0.958) +- (0,0.021)
   (0.97,0.971) +- (0,0.016)
   (0.98,0.976) +- (0,0.015)
   (0.99,0.985) +- (0,0.011)};
\nextgroupplot[title={Compute saved ($\delta = 0$)},
  ymin=0, ymax=62, ylabel={Saved (\%)}]
\addplot[modelQ, error bars/.cd, y dir=both, y explicit] coordinates
  {(0.90,47.4) +- (0,10.1)
   (0.91,46.9) +- (0,11.0)
   (0.92,46.3) +- (0,11.1)
   (0.93,42.8) +- (0,11.7)
   (0.94,42.1) +- (0,11.4)
   (0.95,39.3) +- (0,10.1)
   (0.96,34.7) +- (0,13.5)
   (0.97,28.8) +- (0,13.3)
   (0.98,25.8) +- (0,12.4)
   (0.99,16.7) +- (0,10.5)};
\end{groupplot}
\end{tikzpicture}
  \caption{Unmargined ($\delta = 0$) cascade on TextCraft with
  Qwen-2.5-7B over a
  fine target grid (mean $\pm$ one standard deviation, 20 seeds).
  Left: test recall hugs the diagonal, dipping below it at several
  targets; the margin primarily absorbs per-seed dispersion around the
  mean (Table~\ref{tab:deltaviol}), while mean-level bias is negligible.
  By the shift identity, deploying
  margin $\delta$ at target $\rho^\star$ reads both panels at
  $\rho^\star + \delta$.}
  \label{fig:deltasweep}
\end{figure}

For fixed splits and seeds, the margin rule's feasible set
$\{\mathbf{t} : \hat{\rho}_{\mathrm{val}}(\mathbf{t}) \ge \rho^\star
+ \delta\}$ depends on $(\rho^\star, \delta)$ only through their sum,
so the budget deployed under margin $\delta$ at target $\rho^\star$
is identical, seed by seed, to the unmargined budget at target
$\rho^\star + \delta$. A single unmargined sweep over a fine target
grid therefore characterizes \emph{every} margin size at once:
choosing $\delta$ means choosing where on the curves of
Figure~\ref{fig:deltasweep} to operate, and the cost of a margin is
the savings drop between $\rho^\star$ and $\rho^\star + \delta$.

Table~\ref{tab:deltaviol} reads off the per-seed violation rates.
Unmargined selection violates the floor in up to $12/20$ seeds;
$\delta = 0.01$ helps but still admits up to $8/20$; $\delta = 0.02$
caps violations at $4/20$ at every target, and at $3/20$ for
$\rho^\star \ge 0.91$. Its price follows the
slope of the savings curve: near-zero at loose targets (about one
point of savings at $\rho^\star = 0.90$) and steepest where the
frontier falls fastest (about ten points at $\rho^\star = 0.95$ in
this batch). This sweep uses its own fixed seed stream to characterize
margin sensitivity; the main-body allocation numbers come from a
separate evaluation.

\begin{table}[H]
\centering
\begin{tabular}{@{}lccc@{}}
\toprule
& \multicolumn{3}{c}{Seeds below floor (of 20)} \\
\cmidrule(lr){2-4}
$\rho^\star$ & $\delta = 0$ & $\delta = 0.01$ & $\delta = 0.02$ \\
\midrule
0.90 & 8  & 7 & 4 \\
0.91 & 8  & 6 & 3 \\
0.92 & 7  & 4 & 2 \\
0.93 & 6  & 4 & 1 \\
0.94 & 8  & 4 & 3 \\
0.95 & 10 & 7 & 1 \\
0.96 & 12 & 5 & 3 \\
0.97 & 10 & 8 & 3 \\
\bottomrule
\end{tabular}
\caption{Number of seeds (of 20) whose test recall falls below the
target, per margin size; all columns are derived from the single
unmargined sweep via the shift identity.
$\delta = 0.02$ caps violations at $4/20$ everywhere, and at
$3/20$ for $\rho^\star \ge 0.91$.}
\label{tab:deltaviol}
\end{table}

\suppformatbreak
\subsection*{E. ALFWorld Stress-Test Details}

Table~\ref{tab:alfworld} gives the full per-target results behind the
paper's ALFWorld stress test. The table reports the Qwen
models evaluated with the recorded multi-rollout protocol. The
separate Llama pilot is discussed only as a calibration boundary
because its success population is too small for a meaningful
recall-calibrated cascade evaluation.
The fixed subset contains the first $20$ tasks from each of six
high-level task families ($120$ tasks total), with reset failures
retained in the attempted-rollout counts. In the Llama pilot, a
format-focused retry on a more favorable solved-any subset solves only
$10/68$ tasks ($14.7\%$), reinforcing its role as an out-of-scope,
low-success calibration boundary.

\begin{table}[!htbp]
\centering
\scriptsize
\setlength{\tabcolsep}{2.8pt}
{\renewcommand{\arraystretch}{0.92}
\begin{tabular}{@{}lccccc@{}}
\toprule
Model & Target & Test recall & Cascade & Single & Success/attempts \\
\midrule
Qwen-2.5-7B & 0.90 & $0.891 \pm 0.047$ & $\mathbf{19.7 \pm 5.3}$ & $11.4$ & $250/477$ \\
             & 0.92 & $0.900 \pm 0.052$ & $\mathbf{18.3 \pm 6.8}$ & $7.9$  &  \\
             & 0.95 & $0.943 \pm 0.045$ & $\mathbf{10.7 \pm 6.9}$ & $3.3$  &  \\
             & 0.97 & $0.969 \pm 0.046$ & $\mathbf{5.5 \pm 6.5}$  & $1.1$  &  \\
\midrule
Qwen3-1.7B  & 0.90 & $0.960 \pm 0.076$ & $\mathbf{4.3 \pm 8.1}$  & $3.7$  & $106/477$ \\
             & 0.92 & $0.960 \pm 0.076$ & $\mathbf{4.3 \pm 8.1}$  & $2.0$  &  \\
             & 0.95 & $0.974 \pm 0.062$ & $\mathbf{2.4 \pm 4.9}$  & $0.0$  &  \\
             & 0.97 & $0.974 \pm 0.062$ & $\mathbf{2.4 \pm 4.9}$  & $0.0$  &  \\
\bottomrule
\end{tabular}
}
\caption{ALFWorld stress test (compute saved in \%).
Both models use $477$ recorded rollouts; success/attempts reports the
base success population.}
\label{tab:alfworld}
\end{table}

\subsection*{F. Probe Data Cost and Break-Even Accounting}

The pipeline requires labeled complete trajectories before deployment.
The environment already supplies the binary success label, and the
same logged trajectories are reused for cross-fitting, calibration,
and budget search. In our offline implementation, however, hidden
activations are recovered by a teacher-forced replay, so the one-time
cost includes both collecting the original trajectories and replaying
them for feature extraction. Fitting the linear probes is inexpensive
relative to LLM generation, but trajectory collection can still be
costly, especially for an independent certification set.

Let $C_{\mathrm{data}}$ denote the total one-time cost of trajectory
collection, activation extraction, and probe fitting in a common unit
such as generated-token-equivalents. Let $\bar C$ be the mean baseline
cost per deployed episode, $s$ the expected saved fraction, and
$C_{\mathrm{mon}}$ the per-episode monitoring overhead in the same
unit. The deployment breaks even after
\[
  N_{\mathrm{BE}}
  = \frac{C_{\mathrm{data}}}{s\bar C-C_{\mathrm{mon}}},
  \qquad s\bar C>C_{\mathrm{mon}}.
\]
If the denominator is non-positive, the monitor never repays its data
and systems cost. Existing trajectory logs reduce $C_{\mathrm{data}}$
to incremental extraction and fitting. Certification is separate:
$n_{\mathrm{pos}}$ successes require roughly
$n_{\mathrm{pos}}/p_{\mathrm{succ}}$ attempts at base success rate
$p_{\mathrm{succ}}$. Because serving overhead is unmeasured, token
savings and end-to-end cost should be interpreted separately.

\suppformatbreak
\subsection*{G. Additional Main-Text Diagnostics}

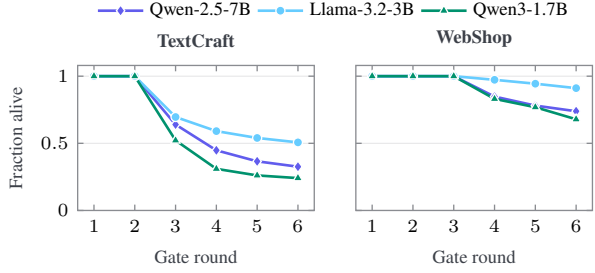
\begin{figure}[H]
  \centering
  \begin{tikzpicture}
\begin{groupplot}[
  paper,
  group style={group size=2 by 1, horizontal sep=0.55cm,
    ylabels at=edge left, yticklabels at=edge left},
  width=0.56\linewidth, height=3.5cm,
  xmin=0.6, xmax=6.4, xtick={1,2,3,4,5,6},
  xlabel={Gate round}, ymin=0, ymax=1.08,
  ylabel={Fraction alive}, legend columns=3]
\nextgroupplot[title={TextCraft}, legend to name=alivelegend]
\addplot[modelQ] coordinates {(1,1.000)(2,1.000)(3,0.640)(4,0.448)(5,0.366)(6,0.326)};
\addplot[modelL] coordinates {(1,1.000)(2,1.000)(3,0.696)(4,0.591)(5,0.540)(6,0.507)};
\addplot[cGreen, line width=1pt, mark=triangle*, mark size=2pt,
  mark options={fill=cGreen, draw=white, line width=0.5pt}] coordinates
  {(1,1.000)(2,1.000)(3,0.522)(4,0.310)(5,0.261)(6,0.241)};
\legend{Qwen-2.5-7B, Llama-3.2-3B, Qwen3-1.7B}
\nextgroupplot[title={WebShop}]
\addplot[modelQ] coordinates {(1,1.000)(2,1.000)(3,1.000)(4,0.848)(5,0.781)(6,0.739)};
\addplot[modelL] coordinates {(1,1.000)(2,1.000)(3,1.000)(4,0.973)(5,0.944)(6,0.911)};
\addplot[cGreen, line width=1pt, mark=triangle*, mark size=2pt,
  mark options={fill=cGreen, draw=white, line width=0.5pt}] coordinates
  {(1,1.000)(2,1.000)(3,1.000)(4,0.831)(5,0.769)(6,0.679)};
\end{groupplot}
\node[anchor=south] at ($(group c1r1.north)!0.5!(group c2r1.north)+(0,0.35cm)$)
  {\pgfplotslegendfromname{alivelegend}};
\end{tikzpicture}
  \caption{Fraction of episodes still running at each gate round in
  the full $2\times3$ matrix. TextCraft episodes often terminate
  early, especially for Qwen3-1.7B, while WebShop episodes remain alive
  longer; this controls how much compute a later gate can save.}
  \label{fig:alive}
\end{figure}

\begin{figure*}[t]
  \centering
  \begin{tikzpicture}
\begin{groupplot}[
  paper,
  group style={group size=3 by 2, horizontal sep=0.42cm, vertical sep=0.82cm,
    ylabels at=edge left, yticklabels at=edge left},
  width=0.34\textwidth, height=3.05cm,
  xmin=0.893, xmax=0.977, xtick={0.90,0.92,0.95,0.97},
  x tick label style={font=\scriptsize, rotate=45, anchor=north east,
    /pgf/number format/.cd, fixed, precision=2},
  ymin=-3, ymax=65, ylabel={Saved (\%)},
  xlabel={Recall target $\rho^\star$},
  legend columns=3]
\nextgroupplot[title={TextCraft Qwen-2.5}, xticklabels=\empty, xlabel={},
  legend to name=frontierlegend]
\addplot[lineP] coordinates {(0.90,60.2)(0.92,56.6)(0.95,45.0)(0.97,31.6)};
\addplot[lineS] coordinates {(0.90,40.1)(0.92,34.8)(0.95,25.8)(0.97,15.9)};
\addplot[lineU] coordinates {(0.90,8.3)(0.92,8.3)(0.95,0.0)(0.97,0.0)};
\legend{Cascade, Single-gate, Uniform}
\nextgroupplot[title={TextCraft Llama}, xticklabels=\empty, xlabel={}]
\addplot[lineP] coordinates {(0.90,34.7)(0.92,25.8)(0.95,17.2)(0.97,6.9)};
\addplot[lineS] coordinates {(0.90,22.0)(0.92,17.1)(0.95,10.7)(0.97,6.0)};
\addplot[lineU] coordinates {(0.90,3.4)(0.92,0.0)(0.95,0.0)(0.97,0.0)};
\nextgroupplot[title={TextCraft Qwen3}, xticklabels=\empty, xlabel={}]
\addplot[lineP] coordinates {(0.90,47.3)(0.92,43.3)(0.95,26.8)(0.97,14.8)};
\addplot[lineS] coordinates {(0.90,29.9)(0.92,23.2)(0.95,14.3)(0.97,7.1)};
\addplot[lineU] coordinates {(0.90,5.9)(0.92,4.6)(0.95,0.0)(0.97,0.0)};
\nextgroupplot[title={WebShop Qwen-2.5}]
\addplot[lineP] coordinates {(0.90,35.9)(0.92,31.8)(0.95,20.6)(0.97,11.3)};
\addplot[lineS] coordinates {(0.90,19.2)(0.92,14.4)(0.95,8.0)(0.97,5.0)};
\addplot[lineU] coordinates {(0.90,0.0)(0.92,0.0)(0.95,0.0)(0.97,0.0)};
\nextgroupplot[title={WebShop Llama}]
\addplot[lineP] coordinates {(0.90,24.7)(0.92,20.0)(0.95,13.6)(0.97,12.0)};
\addplot[lineS] coordinates {(0.90,2.8)(0.92,2.1)(0.95,1.1)(0.97,0.0)};
\addplot[lineU] coordinates {(0.90,0.0)(0.92,0.0)(0.95,0.0)(0.97,0.0)};
\nextgroupplot[title={WebShop Qwen3}]
\addplot[lineP] coordinates {(0.90,54.9)(0.92,52.3)(0.95,41.5)(0.97,26.4)};
\addplot[lineS] coordinates {(0.90,8.3)(0.92,6.1)(0.95,3.8)(0.97,1.8)};
\addplot[lineU] coordinates {(0.90,0.0)(0.92,0.0)(0.95,0.0)(0.97,0.0)};
\end{groupplot}
\node[anchor=south] at ($(group c1r1.north)!0.5!(group c3r1.north)+(0,0.35cm)$)
  {\pgfplotslegendfromname{frontierlegend}};
\end{tikzpicture}
  \caption{Compute savings versus global recall target for all three
  allocations and matrix cells. The cascade dominates
  single-gate and uniform allocation in every configuration.}
  \label{fig:frontier}
\end{figure*}
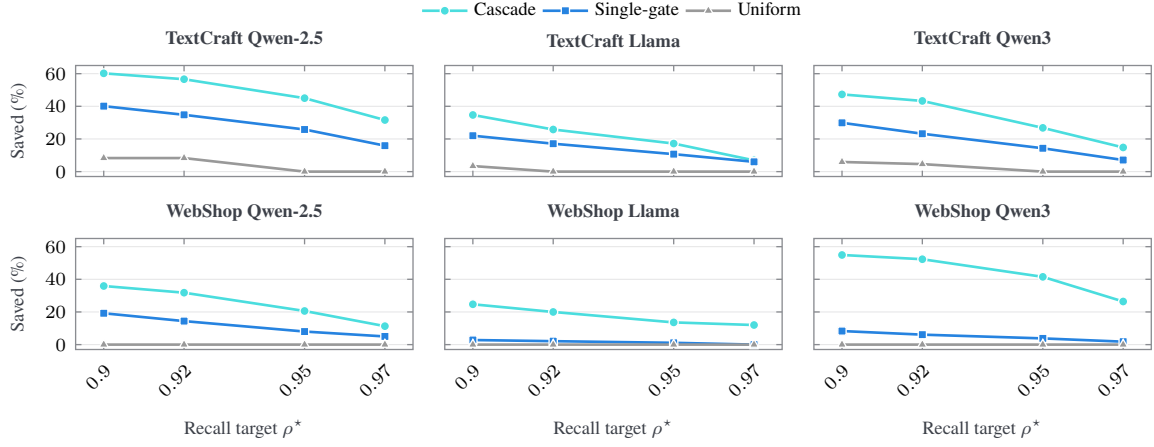

The following plots preserve two complete diagnostics omitted from the
seven-page technical narrative for space. Figure~\ref{fig:alive} shows
how much future compute remains at each candidate gate, while
Figure~\ref{fig:frontier} expands the main paper's numerical allocation
comparison across every recall target.

\suppformatbreak
\subsection*{H. Additional Related Work}

\paragraph{Predicting success from internal representations.}
Frozen-activation probes decode knowledge, truth, and hallucination
signals that models do not reliably verbalize
\citep{alain2016understanding,belinkov2022probing,kadavath2022language,
azaria2023internal,burns2023discovering,marks2023geometry,
orgad2025llms,kossen2024semantic,ji2024llm}. Internal states also
anticipate chain-of-thought success and encode self-verification
\citep{afzal2025knowing,zhang2025reasoning}. In agent settings,
\citet{anon2026conformal_interp} recover success/failure directions for
representation steering, while MEMPROBE targets persistent memory
rather than online residual activations \citep{ma2026memprobe}, and
\citet{mehta2026premature} study commitment, deliberately separated
from correctness. These results establish that outcome-relevant signal
exists in the residual stream and is linearly accessible; we ask when
it becomes actionable and turn it into a stopping rule with controlled
error.

\paragraph{Failure detection and monitoring of LLM agents.}
Decisive agent errors often arise early and compound
\citep{deng2025agentdebug,anon2026observability}; AgentRx localizes such
steps post hoc \citep{barke2026agentrx}. Online approaches range from
text or embedding auditors, including AgentForesight, weakly supervised
alerters, and MASC \citep{zhang2026foresight,anon2026alphastop,
shen2025masc}, to behavioral statistics and low-cost runtime signals
\citep{chen2026signals,pham2026agentstop}. Richer monitors use
diagnostic agents, typed events, or execution graphs
\citep{mittapalli2026trace,dellibarda2026poirot,
huang2026prefixguard,zhao2026grade}; other methods alter refusal,
stop on repetition, or elicit self-reports
\citep{guan2026futile,shrivastava2026semantic,anon2026bagen,
jiang2025runaway}. Our activation reader needs no auxiliary LLM pass,
and its cascade controls retained-success recall over all checks.

\paragraph{Conformal prediction and risk control for LLMs.}
Split conformal and related calibration give finite-sample guarantees
under exchangeability
\citep{vovk2005algorithmic,papadopoulos2002inductive,
lei2018distribution,angelopoulos2023gentle}, with extensions to general
risks \citep{bates2021distribution,angelopoulos2025learn,
angelopoulos2024conformal}. LLM applications include prediction sets,
factuality filters, and help-seeking planners
\citep{quach2024conformal,mohri2024language,cherian2024large,
ren2023robots}. \citet{davidov2026quit} derive optimal quitting from a
learned value function. CALM composes token exits under a sequence-level
constraint \citep{schuster2022confident}, while FIPER calibrates
robot-policy alarms from successful rollouts \citep{romer2025fiper}.
Our fixed-policy certificate instead controls retained-success recall
across an episode-level sequence of gates.

\paragraph{Adaptive allocation of inference compute.}
RLVR and recovery training optimize sampled trajectories from task
outcomes \citep{shao2024deepseekmath,deepseekai2025r1,
qu2024recursive}. At inference time, adaptive self-consistency stops
once answers stabilize
\citep{aggarwal2023lets,li2024escape,manvi2024adaptive}, model cascades
route by difficulty \citep{chen2023frugalgpt,zellinger2025cost}, and
overthinking methods stop redundant reasoning
\citep{anon2025refrain,mao2025escot}. Atropos predicts ongoing reasoning
success \citep{kim2026atropos}; early-exit systems vary decoder depth
\citep{chen2023eellm,miao2024efficient}; and process reward models
prune partial trajectories
\citep{lightman2024lets,zhang2024rest,xia2025agentrm}. We instead decide
whether one running agent episode is worth finishing and pair savings
with calibrated retained-success recall.

\end{document}